\documentclass[runningheads]{llncs}
\usepackage{graphicx}
\usepackage{comment}
\usepackage{amsmath,amssymb} 
\usepackage{color}
\usepackage{subfigure}
\usepackage{multirow}
\usepackage{multicol}
\graphicspath{{pics/}}
\begin{document}
\title{A Benchmark of Ocular Disease Intelligent Recognition: One Shot for Multi-disease Detection}
\titlerunning{A Benchmark of Ocular Disease Intelligent Recognition}
%
\author{Ning Li\inst{1}
\and Tao Li\inst{1,2}
\and Chunyu Hu\inst{1}
\and Kai Wang\inst{1}
\and Hong Kang\inst{1,3}}
\authorrunning{N Li et al.}
%
\institute{
College of Computer Science, Nankai university, Tianjin, 300350, China \and
State Key Laboratory of Computer Architecture, Institute of Computing Technology, Chinese Academy of Science, Beijing, 100190, China \and
Beijing Shanggong Medical Technology Co. Ltd, Beijing, 100176, China\\
\email{kanghong@nankai.edu.cn}
}
\maketitle              
\begin{abstract}
In ophthalmology, early fundus screening is an economic and effective way to prevent blindness caused by ophthalmic diseases. Clinically, due to the lack of medical resources, manual diagnosis is time-consuming and may delay the condition. With the development of deep learning, some researches on ophthalmic diseases have achieved good results, however, most of them are just based on one disease. During fundus screening, ophthalmologists usually give diagnoses of multi-disease on binocular fundus image, so we release a dataset with 8 diseases to meet the real medical scene, which contains 10,000 fundus images from both eyes of 5,000 patients. We did some benchmark experiments on it through some state-of-the-art deep neural networks. We found simply increasing the scale of network cannot bring good results for multi-disease classification, and a well-structured feature fusion method combines characteristics of multi-disease is needed. Through this work, we hope to advance the research of related fields.
\keywords{fundus images  \and multi-disease \and image classification \and neural network \and computer-aided diagnosis.}
\end{abstract}
\section{Introduction}
Fundus diseases are the leading causes of blindness among human beings worldwide~\cite{quigley2006number}. Among them, diabetic retinopathy (DR), glaucoma, cataract and age-related macular degeneration (AMD) are the most popular ophthalmic diseases.
According to related surveys,
there will be over 400 million people with DR by 2030, and the glaucoma patients will reach 80 million people worldwide by 2020~\cite{costagliola2009pharmacotherapy}.
These ophthalmic diseases have become a serious public health problem in the world.
Most importantly, the ophthalmic disease has irreversible properties, which could cause unrecoverable blindness.
In clinical scenarios, early detection of these diseases can prevent visual impairment.
However, the number of ophthalmologists are out of balance compared with that of patients seriously.
Moreover, fundus screening manually is time-consuming and relies heavily on ophthalmologists' experience.
These reasons make it difficult to perform large-scale fundus screening.
Therefore, an automatic computer-aided diagnosis algorithm for ophthalmic diseases screening is important particularly.

However, designing such an effective computer-aided diagnosis algorithm is challenging.
For example, microaneurysm is an important reference for DR screening~\cite{li2019diagnostic}. However, the size of microaneurysm is very small, so that it is hard to be detected and it is easy to be confused with other lesions.
Meantime, the low contrast between the lesion pixels and the background pixels, the irregular shape of the lesions and the large differences between the same lesion points caused by different cameras also make it difficult to accurately identify ophthalmic diseases.

Although there have already been some deep learning models for ophthalmic disease screening, and they achieve remarkable performance.
We found there are some limitations.
1) \textit{Single disease}. Most of the identification models only concentrate on one ophthalmic disease~\cite{tan2018age}~\cite{zhang2019automatic}, and most of the dataset they use provide annotations for only one kind of ophthalmic disease. However, considering the actual needs of patients with fundus disease in daily life, we believe that establishing a more effective and comprehensive fundus screening system that can detect multiple diseases is necessary.
2) \textit{Single eye}. Existing datasets are based on a fundus image~\cite{choi2017multi}~\cite{chen2014multiple}, but in real clinical scenarios, ophthalmologists usually diagnose patients with information from both eyes.

To solve the problem above-mentationed, in this paper, we release a publicly available dataset for multiple ophthalmic diseases detection. Different from the current published international monocular dataset, our dataset contains 5,000 pairs of binocular images, i.e. 10,000 images in total.
Moreover, we provide annotations for 8 diseases on binocular images,
which means that there may be multiple diseases for one patient.
In addition, multi-disease screening is more complicated than the current single disease screening, and there are few relevant studies to learn from. Hence, we performed experiments on several popular deep learning based classification networks for multi-disease classification.
Extensive experiments show simply increasing the scale of network cannot lead to performance improvement, and a well-structured feature fusion method which combines characteristics of multiple diseases is needed.

In summary, the main contributions of our work are as follows. First, we collected, annotated and release a multi-disease fundus image dataset named Ophthalmic Image Analysis-Ocular Disease Intelligent Recognition (OIA-ODIR). Second, we performed experiments on nine popular deep neural network models on this dataset, thereby establishing a benchmark. At last, by presenting this dataset, we hope that it can promote the further development of the clinical multi-disease classification research.

\section{Related Work}
At present, the works of ocular disease screening are mainly performed on optical coherence tomography (OCT) images and fundus images. With the development of artificial intelligence in the field of medical image processing, some related methods have achieved pleasing results.

\subsection{OCT Images Recognition}
OCT is one of the commonly used methods for fundus disease examination. After investigation, about 5.35 million OCTs were used in the United States in 2014~\cite{alqudah2020aoct}. OCT has been widely used in clinical because of the advantages of low-loss, high-resolution, noninvasive medical imaging, compared to other methods. Ophthalmologists can observe the patient's retina layers through OCT images, measure these layers, and find minor early fundus lesions, then provide the corresponding treatment~\cite{Ting577}.

At present, the focus of some works is done through OCT image recognition, including segmentation~\cite{lee2009segmentation}, detection~\cite{schlegl2018fully} and classification~\cite{rasti2017macular}, etc. He et al.~\cite{he2019fully} proposed a new way for retina OCT layer surface and lesion segmentation without handcrafted graph.  A novel method for multiclass drusen segmentation in retinal OCT images was proposed by Asgari et al.~\cite{asgari2019multiclass}. Their method consistently outperforms several benchmarks in some ways by using a decoder for each target category and an additional decoder for the areas among the target layers. Marzieh Mokhtari et al.~\cite{mokhtari2019local} calculate local cup to disc ratio by fusing fundus images and OCT B-scans to get the symmetry of two eyes, which can detect early ocular diseases better. Mehta et al.~\cite{mehta2018multilabel} proposed a OCT images system for multi-class, multi-label classification, which augments data by using patient information, such as age, gender and visual acuity data.

Although the work of OCT images for ocular diseases screening has been quite mature, the existing public datasets of OCT images are quite few. In addition, comparing with color fundus images, OCT images have higher requirements on acquisition equipment and are more difficult to obtain. In this paper, we provide a new large-scale color fundus image dataset to encourage further research which could be applied in real clinical scenes.

\subsection{Fundus Images Recognition}
Some related results have been published on fundus image classification. In order not to delay the treatment of patients and to solve the quality classification of fundus images, Zhang et al. proposed an improved residual dense block convolutional neural network to effectively divide fundus images into "good quality" and "poor quality"~\cite{zhang2020automated}. Zhang et al. described a six-level cataract grading method focusing on multi-feature fusion, which extracted features from residual network (ResNet-18) and gray level co-occurrence matrix (GLCM), the results show advanced performance~\cite{zhang2019automatic}. Hong et al. developed a 14-layers deep CNN model that can accurately diagnose diseases in the early stages of AMD and help ophthalmologists perform ocular screening~\cite{tan2018age}.

For multiple diseases recognition. Choi et al. used neural networks and random forest to study classification of 10 fundus diseases on STructured Analysis of the REtina (STARE) database~\cite{choi2017multi}. Chelaramani at al. conducted three tasks on a private dataset, including four common diseases classification, 320 fine-grained classification and generated text diagnosis~\cite{chelaramani2019multi}. On a public database named Singapore Malay Eye Study
(SiMES)~\cite{foong2007rationale}, Chen et al. performed multi-label classification of three common fundus diseases on an image, with Entropic Graph regularized Probabilistic Multi-label learning~\cite{chen2014multiple}.

With the rapid development of artificial intelligence, computer-assisted fundus disease diagnosis have gradually developed. Although the above works have obtained quite good results, they still have limitations to a certain extent. Many of their works are based on a single disease solely, when other fundus diseases need to be detected, the structure need to be redesigned, which undoubtedly makes research cumbersome and inefficient. At the same time, in the process of image processing, some works requires artificial designed features, which makes the operation cumbersome and requires a lot of human prior knowledge.

Currently, many existing datasets for the research on fundus diseases are either too small or the types of diseases are too single, which makes them extremely difficult to apply their work to practical clinical scenarios. Although there are already some fundus datasets for multi-disease research, they are relatively few in number of images and types of diseases. In clinical diagnosis, patients usually have more than one ocular disease, so it is necessary to publish a large-scale fundus image dataset containing multiple diseases.

\section{Our Dataset}
As the best of our knowledge, there are few publicly available fundus image datasets with multi-disease annotations on one image.
However, in clinical application, more than one prediction could be given when observing binocular fundus image.
At the same time, ophthalmologists make a diagnosis based on the patient's fundus image, age and other information.
This has prompted us to collect and release a multi-modal fundus dataset containing multiple diseases.
In this section, we will introduce some details of our dataset.

\subsection{Image Collection and Annotation}
\subsubsection{Collection}
The images of OIA-ODIR are derived from our private clinical fundus databases. The database contains more than 1.6 million images totally, and the fundus images are collected from 487 clinical hospitals in 26 provinces across China.
Each image contains abnormalities in different areas of the fundus caused by various diseases, these areas include macula, optic cup, optic disc, blood vessels, and the entire fundus background. We conducted statistics and analysis on these fundus disease categories, and unified a more detailed classification of a certain disease category into a category to label each image, i.e., the DR of stage 1,2,3,4 are unified as DR et al.

In order to ensure the high-quality of fundus images, we cleaned the images of our private database by filtering out duplicate images and low-quality images. Then, we selected some samples from the remaining fundus images at a suitable ratio for training and testing. Finally, we obtained 10,000 fundus images with 8 types of annotations from the left and right eyes of 5,000 patients. The labeled categories include normal, diabetic retinopathy (DR), glaucoma, cataract, age-related macular degeneration (AMD), hypertension, myopia, and other diseases.

\subsubsection{Annotation}
\begin{figure}[t]
\centering
\subfigure[]{
\begin{minipage}[t]{0.16\linewidth}
\centering
\includegraphics[width=0.7in]{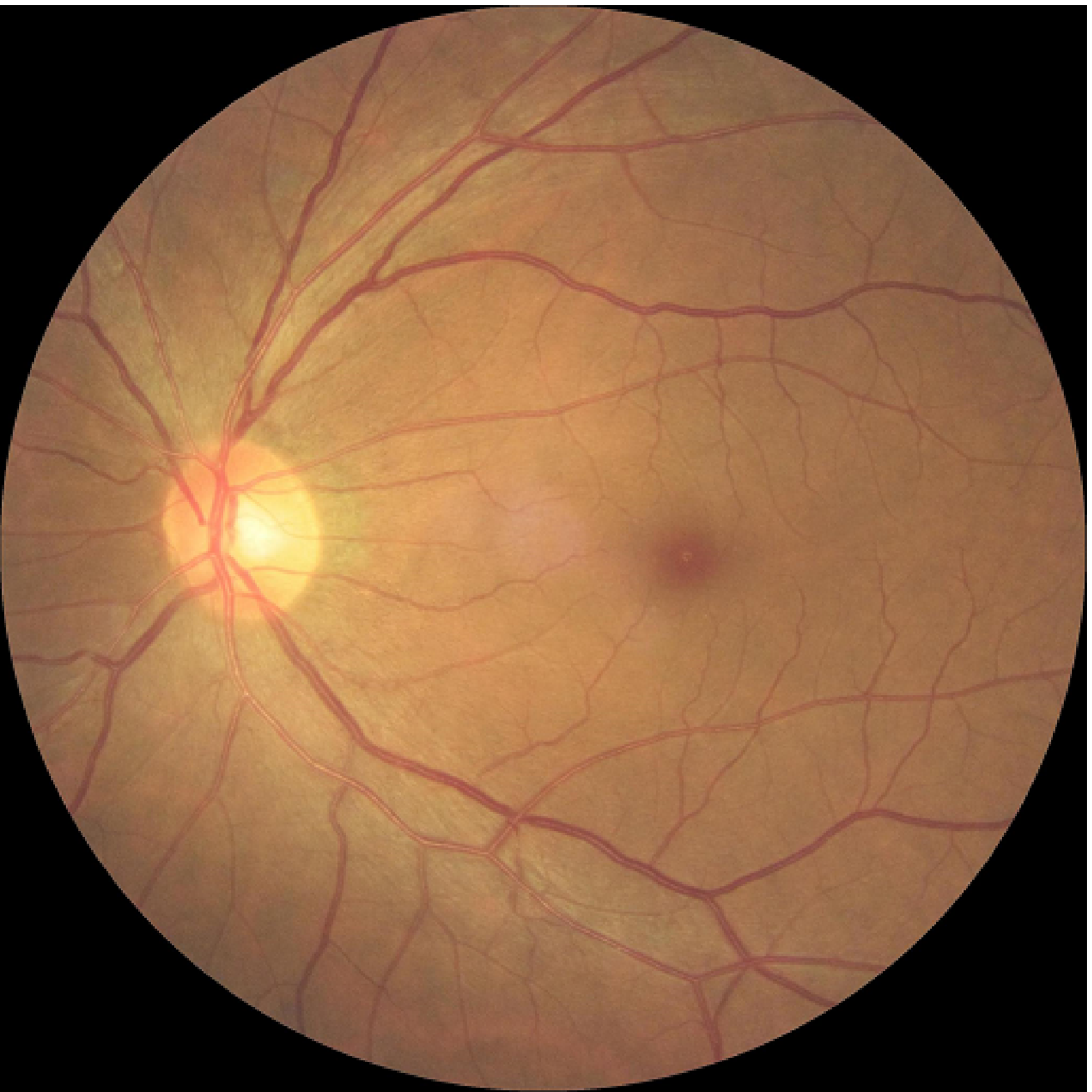} \\
\vspace{0.05in}
\includegraphics[width=0.7in]{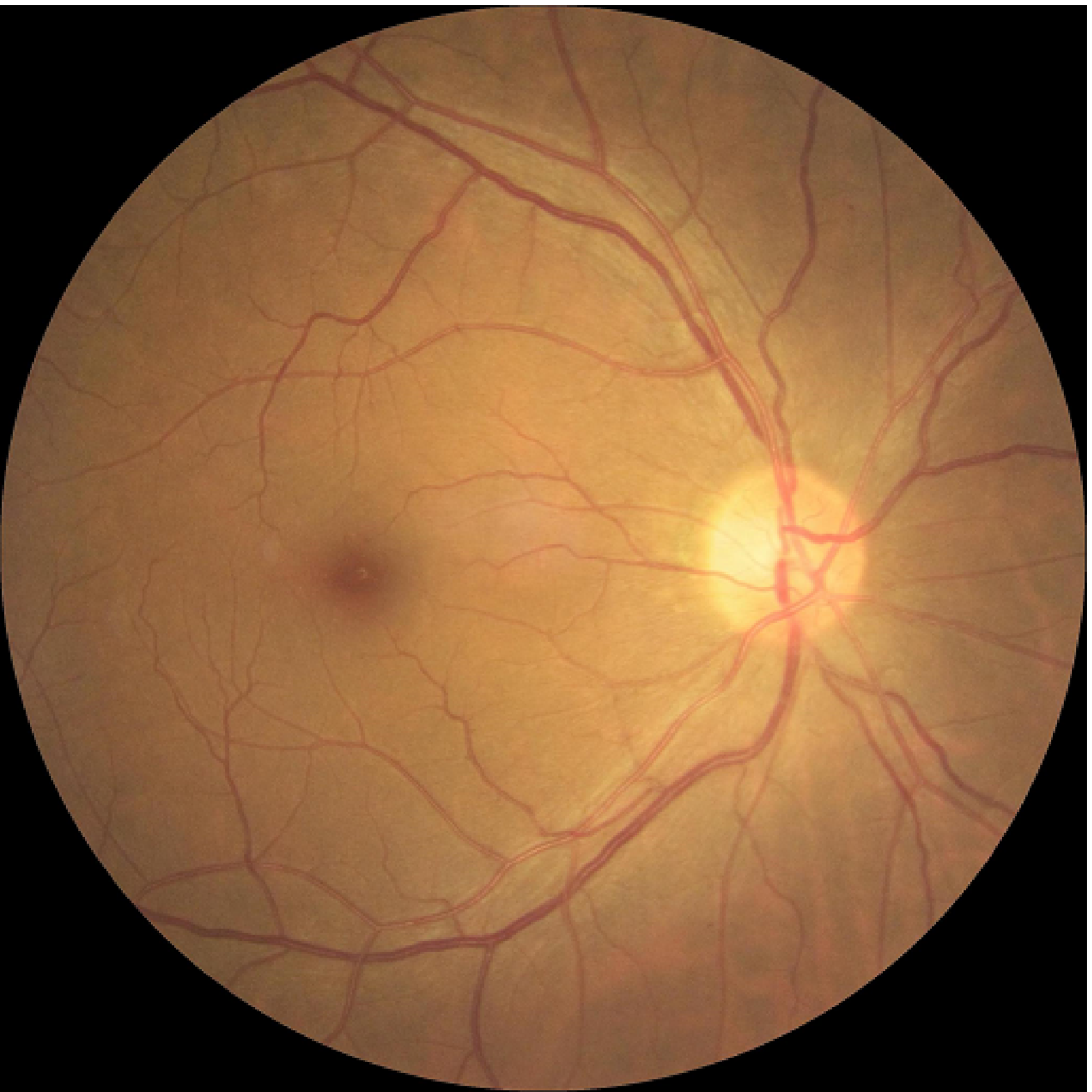}
\end{minipage}
}
\hspace{-0.2in}
\subfigure[]{
\begin{minipage}[t]{0.16\linewidth}
\centering
\includegraphics[width=0.7in]{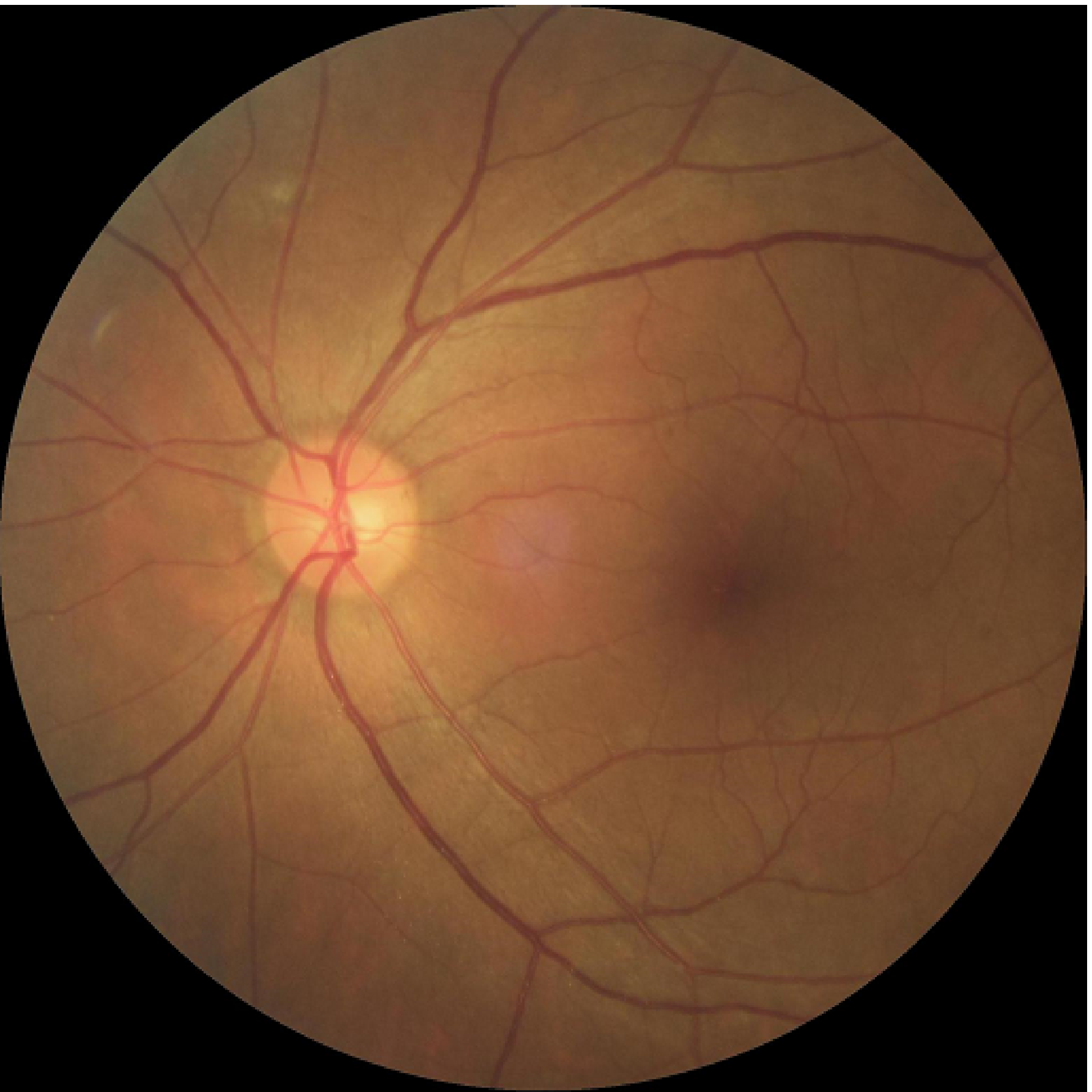} \\
\vspace{0.05in}
\includegraphics[width=0.7in]{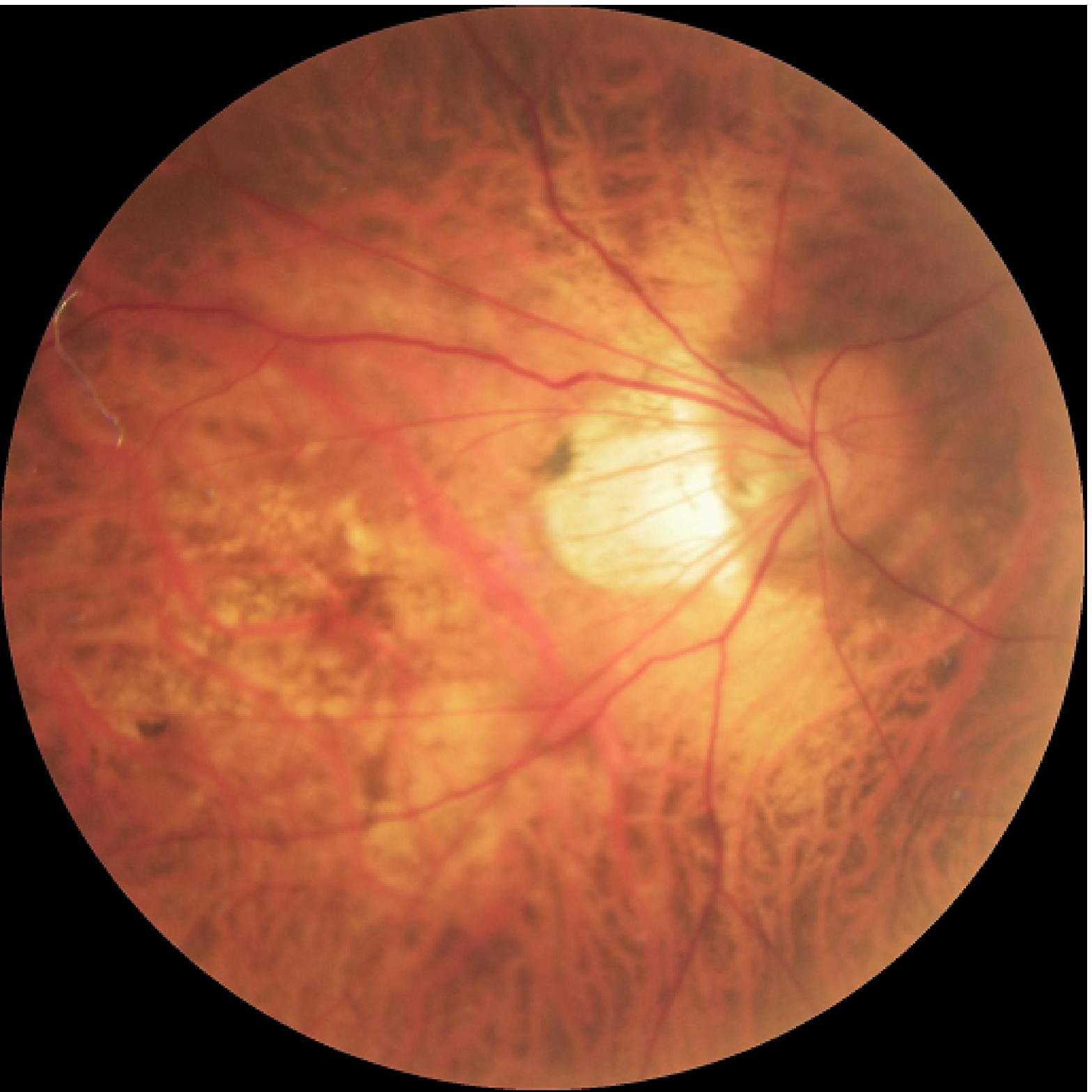}
\end{minipage}
}
\hspace{-0.2in}
\subfigure[]{
\begin{minipage}[t]{0.16\linewidth}
\centering
\includegraphics[width=0.7in]{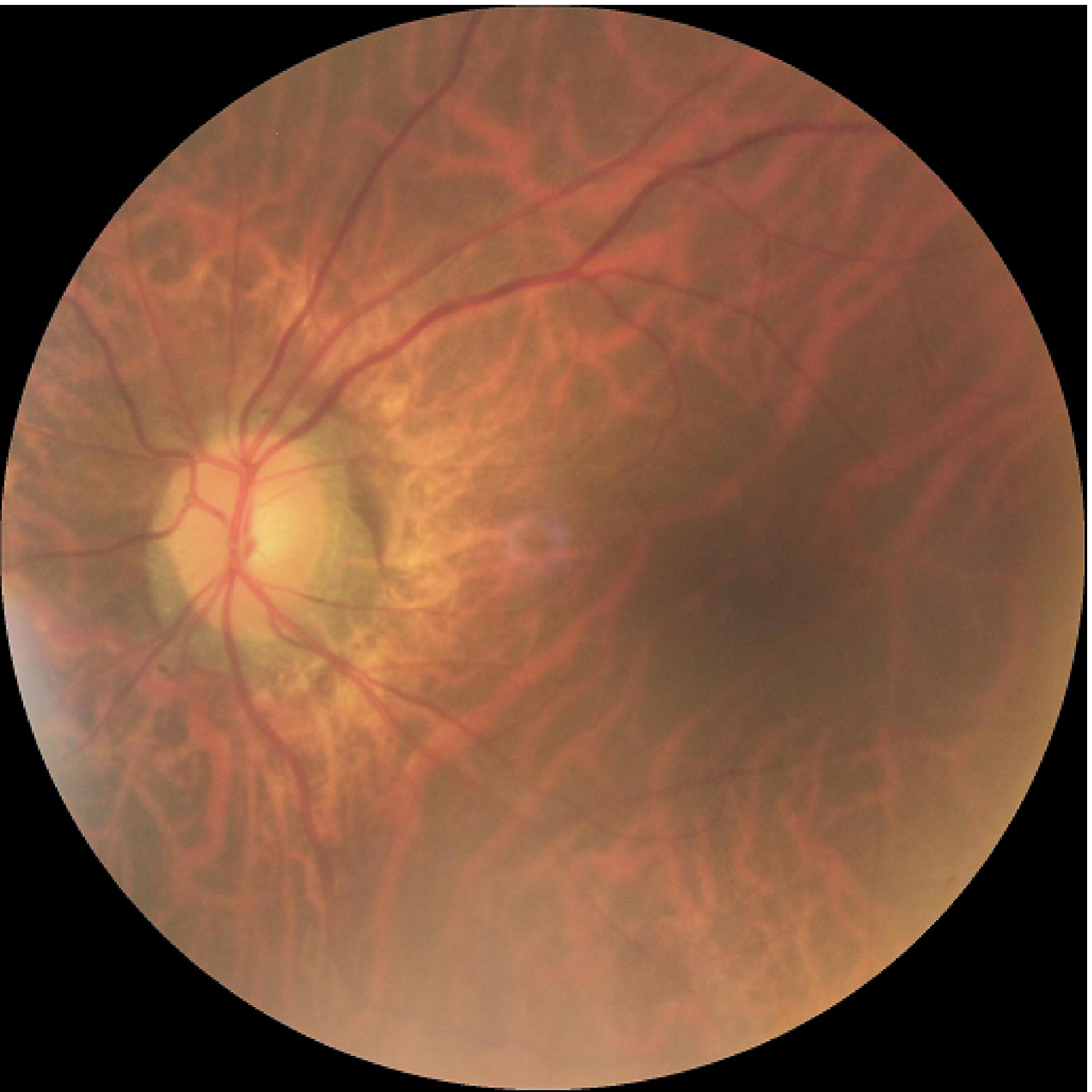} \\
\vspace{0.05in}
\includegraphics[width=0.7in]{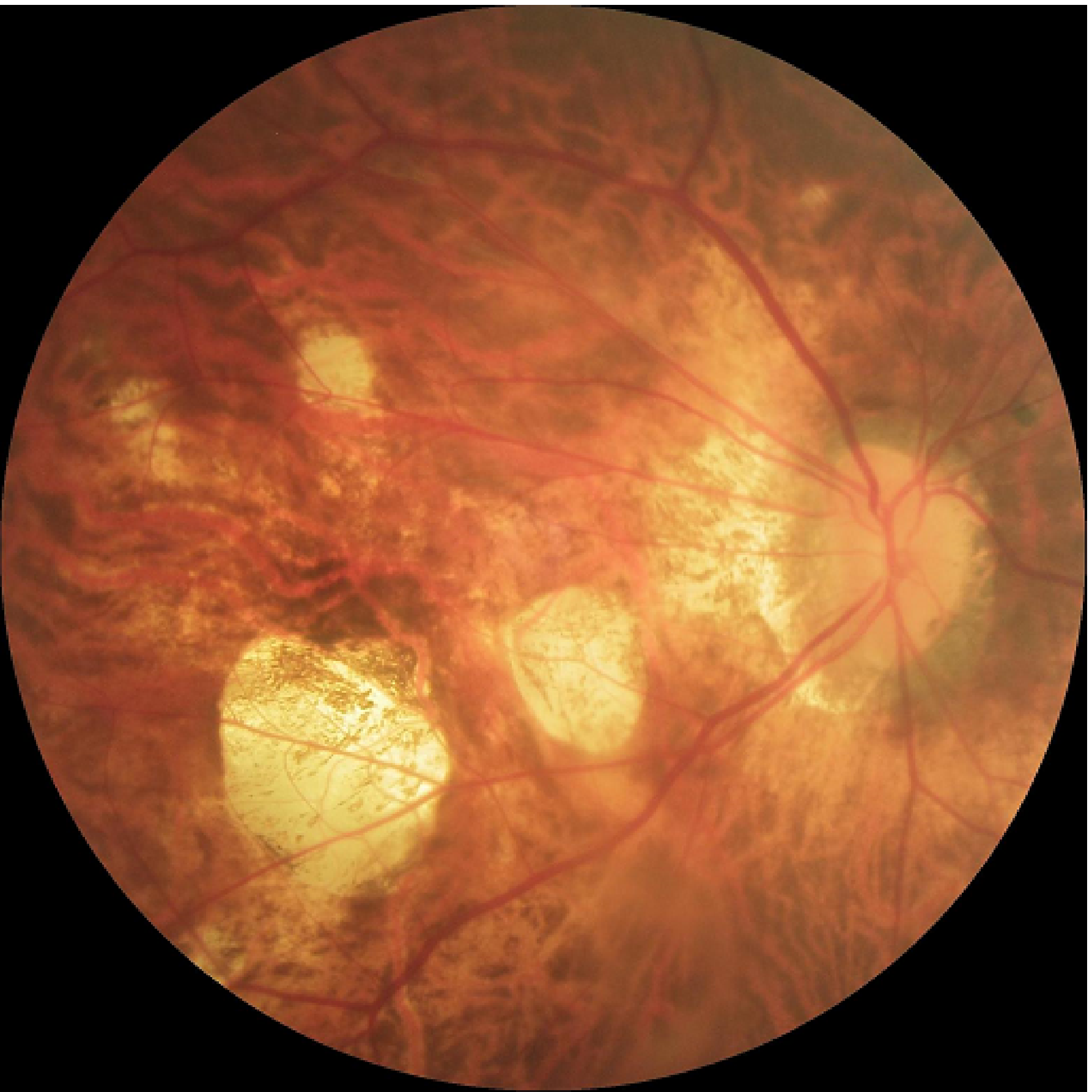}
\end{minipage}
}
\hspace{-0.2in}
\subfigure[]{
\begin{minipage}[t]{0.16\linewidth}
\centering
\includegraphics[width=0.7in]{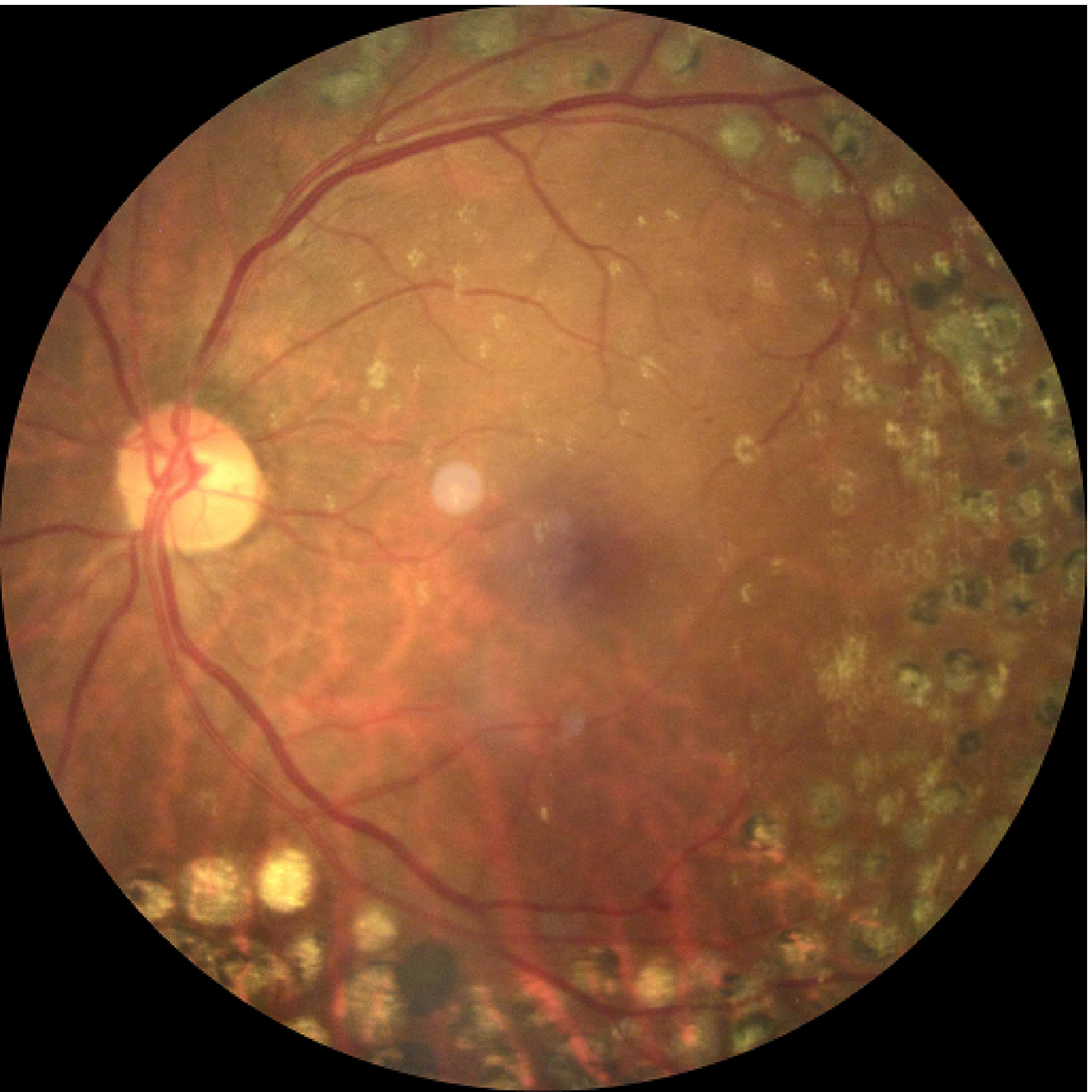} \\
\vspace{0.05in}
\includegraphics[width=0.7in]{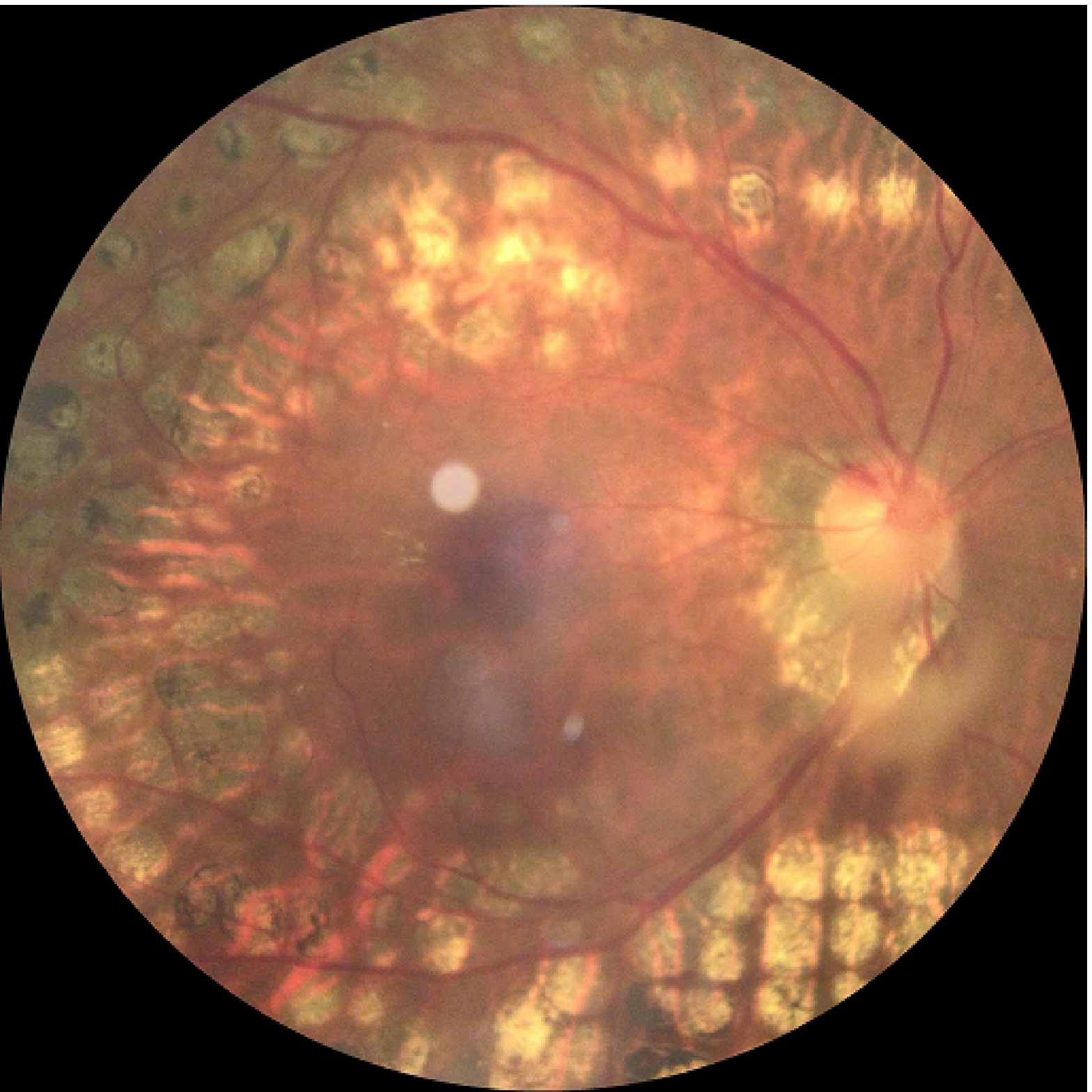}
\end{minipage}
}
\hspace{-0.2in}
\subfigure[]{
\begin{minipage}[t]{0.16\linewidth}
\centering
\includegraphics[width=0.7in]{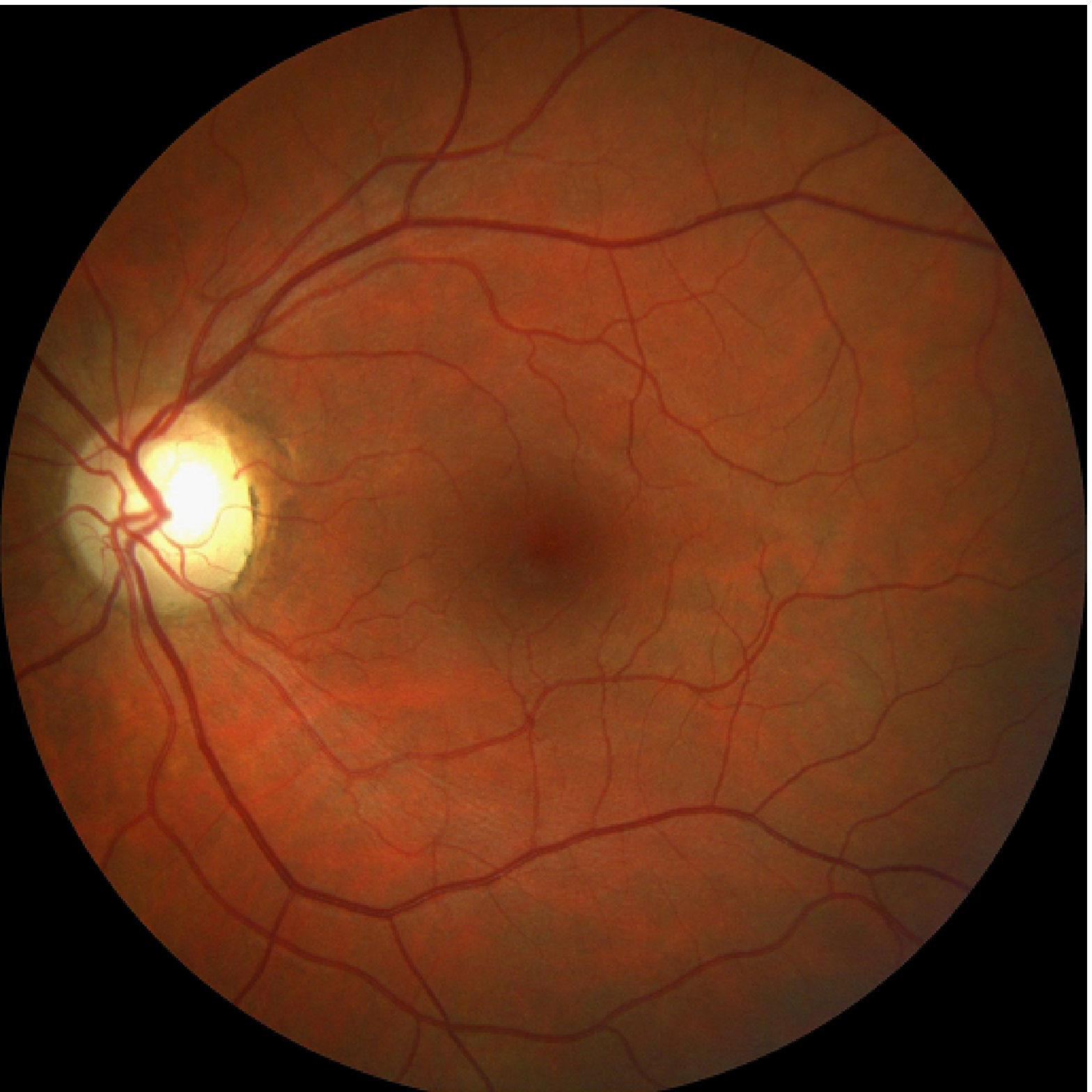} \\
\vspace{0.05in}
\includegraphics[width=0.7in]{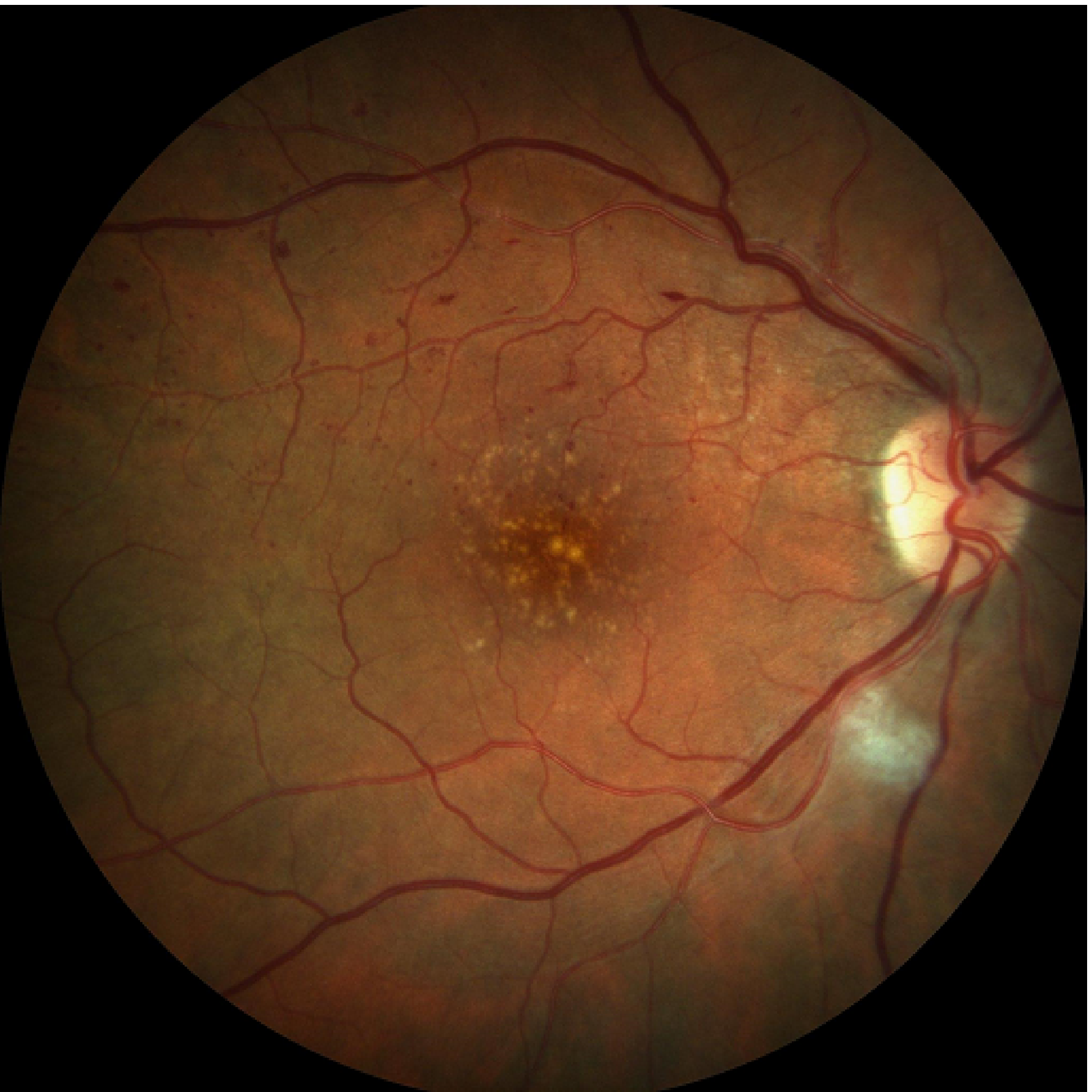}
\end{minipage}
}
\hspace{-0.2in}
\subfigure[]{
\begin{minipage}[t]{0.16\linewidth}
\centering
\includegraphics[width=0.7in]{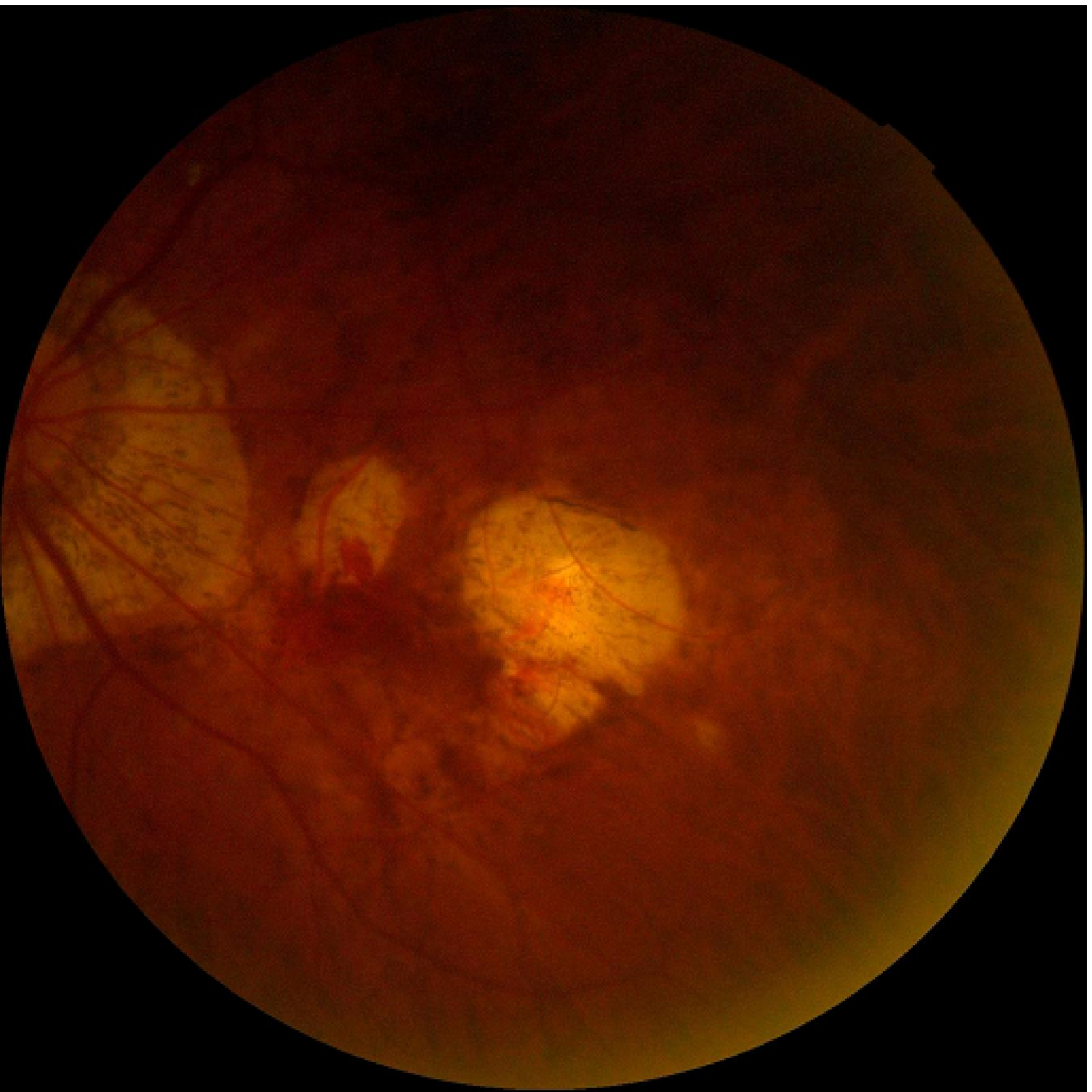} \\
\vspace{0.05in}
\includegraphics[width=0.7in]{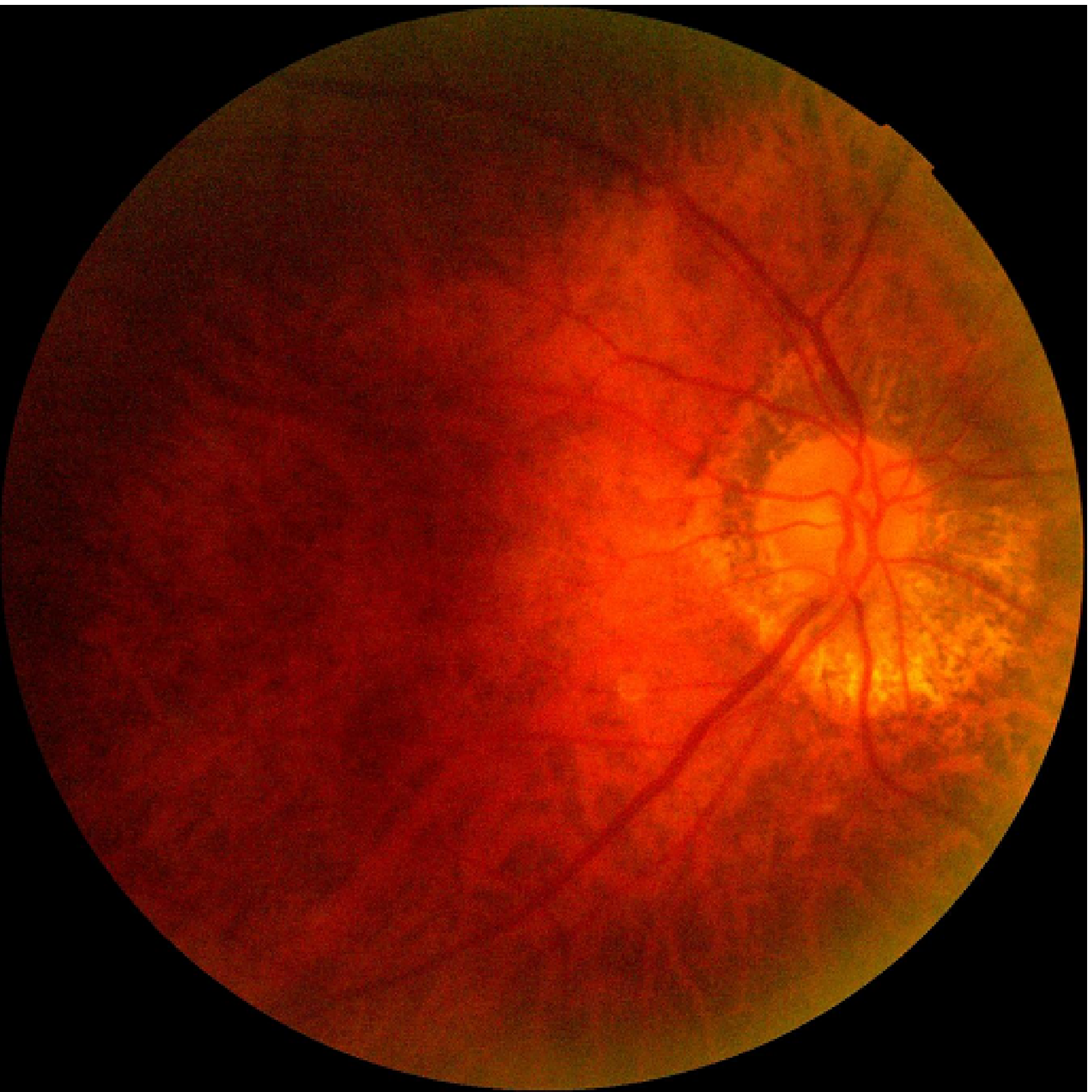}
\end{minipage}
}
\caption{Examples of images in left(top) and right(bottom) eyes. The diagnostic keywords of each column are provided as [left; right]. (a) normal; normal (b) normal; pathological myopia (c) DR; pathological myopia (d) laser spot, vitreous denaturation, DR; laser spot, DR (e) normal; AMD, DR (f) DR, pathological myopia; pathological myopia}
\label{fig2}
\end{figure}

The annotation work of our dataset is done by the professional annotation staff and arbitration team, which took about 10 months to complete. The annotation staff and arbitration team are composed of three ophthalmologists with more than 2 years of clinical experience and three ophthalmologists with more than 10 years of clinical experience in ophthalmology. We strictly follow the corresponding standards and procedures in the process of data annotation. First, three annotators respectively annotate the same batch of fundus images and record the results. If there is any disagreement among the three annotators, the arbitration team will arbitrate the results, the final annotation result shall be based on the consensus of two or more experts. All of these guarantee the persuasiveness of our dataset. At last, some images from our dataset can be seen in Fig.~\ref{fig2}.

\subsection{Features of Dataset}
Currently, there are  already some datasets for ophthalmic disease research. For example, in~\cite{li2019diagnostic}, Gao et al. exposed a fundus image dataset called DDR, which can perform three tasks of image classification, semantic segmentation and object detection. Kaggle DR~\cite{kaggleDR} is the largest fundus image dataset currently used for the classification of DR. Kora~\cite{brandl2016features} is widely used in AMD detection, which contains fundus images of 2840 patients. STARE~\cite{hoover2000locating} has 397 images and can be used for 14 diseases' classification. The Messidor-2 dataset contains 1,200 fundus images for classification of DR and macular degeneration~\cite{decenciere2014feedback}. A dataset related to glaucoma detection research, named ORIGA, contains 650 fundus images, but the dataset is not available~\cite{zhang2010origa}. In addition, there is an unpublished dataset RIM-ONE about optic nerve head segmentation~\cite{fumero2011rim}. While e-optha is public, it can only be used for lesion detection~\cite{decenciere2013teleophta}. SiMES~\cite{foong2007rationale} contains 3150 images of 6 kinds of fundus abnormalities, which can be used for multi-label classification. All these datasets have greatly promoted the development of medical image processing and are of innovative significance. Table~\ref{tab1} shows some statistics for these datasets.

\setlength{\tabcolsep}{4pt}
\begin{table}
\caption{Statistics of the existing ophthalmic disease datasets}
\label{tab1}
\centering
\begin{tabular}{c c c c c c}
\hline
Dataset & Annotations & Images & Multi-disease & Multi-label & Available \\
\hline
DDR~\cite{li2019diagnostic} & DR staging & 13,673 & N & N & Y\\
Kaggle DR~\cite{kaggleDR} & DR staging & 88,702 & N & N & Y\\
KORA~\cite{brandl2016features} & AMD & - & N & N & Y\\
Messidor-2~\cite{decenciere2014feedback} & DR staging, AMD & 1,200 & N & N & Y\\
ORIGA~\cite{zhang2010origa} & glaucoma detection & 650 & N & N & N\\
RIM-ONE~\cite{fumero2011rim} & ONH segementation & 783 & N & N & N\\
e-optha~\cite{decenciere2013teleophta} & lesion detection & 463 & N & N & Y\\
STARE~\cite{hoover2000locating} & 14-disease & 397 & N & N & Y\\
~\cite{wang2019retinal} & 36-disease & - & Y & Y & N\\
SiMES~\cite{foong2007rationale} & 6-disease & 3,150 & Y & Y & Y\\
\hline
ODIR & 8-disease & 10,000 & Y & Y & Y\\
\hline
\end{tabular}
\end{table}


Although many datasets as above have been proposed for ophthalmic disease research, few of them are used for the detection of multiple ophthalmic diseases on one eye, which undoubtedly causes obstacles to the related work for clinical application. As far as we know, our dataset, named OIA-ODIR, is the first internationally launched large-scale multi-type diseases detection dataset based on binocular fundus image. Compared with other fundus image datasets in the same field, our dataset has significant features described as follows.

\begin{enumerate}
\item \textit{Multi-disease:} Unlike most existing fundus image datasets, which only focus on one ophthalmic disease, our dataset contains multiple ophthalmic diseases. As shown in Fig.~\ref{fig3}, these diseases include abnormalities with lesions in different areas of the fundus. According to the International Classification Standard ICO~\cite{muqit2016ico}, DR is divided into four stages. The early stage of DR is characterized by various abnormalities on the retina. For example, there are lesions such as hard exudate, soft exudate, bleeding and neovascularization on the fundus image. Compared with advanced DR, the early stage is not serious and the clinical treatment is significant~\cite{sengupta2019ophthalmic}. For glaucoma, the ophthalmologist usually calculates the ratio of the optic cup to the optic disc. When the ratio is greater than 0.5, the patient is judged to have glaucoma. In recent years, ophthalmologists have also diagnosed by the neuroretinal rim loss, the visual field and the defect of the retinal nerve fiber layer. Clinically, protocols such as the American Cooperative Cataract Research Group (CCRG) method are generally used to classify cataracts. Experienced ophthalmologists compare the fundus images of patients with standard cataract photos to determine the severity of cataract patients~\cite{zhou2019automatic}. Nowadays, AMD is also a common cause of blindness in people of an advanced age, and it is closely related to drusen in the macular area~\cite{garcia2019machine}. Ophthalmologists diagnose the severity of AMD by the size and number of drusen. In the fundus image of a patient with hypertension, we can see that the arteriovenous diameter has a larger ratio. For pathological myopia, there are clear leopard print shapes in the patient's fundus area.

    In the field of ophthalmology research, most of the existing datasets are based on a fundus disease, which makes it difficult to apply related work based on them to the detection of other diseases. Clinically, ophthalmologists may give diagnosis results of various diseases by observing one fundus images. Therefore, as a dataset containing multiple fundus diseases, our dataset is closer to clinical application scenarios.

\begin{figure}[tbp]
\centering
\subfigure[N]{
\begin{minipage}[t]{0.125\linewidth}
\centering
\includegraphics[width=0.5in]{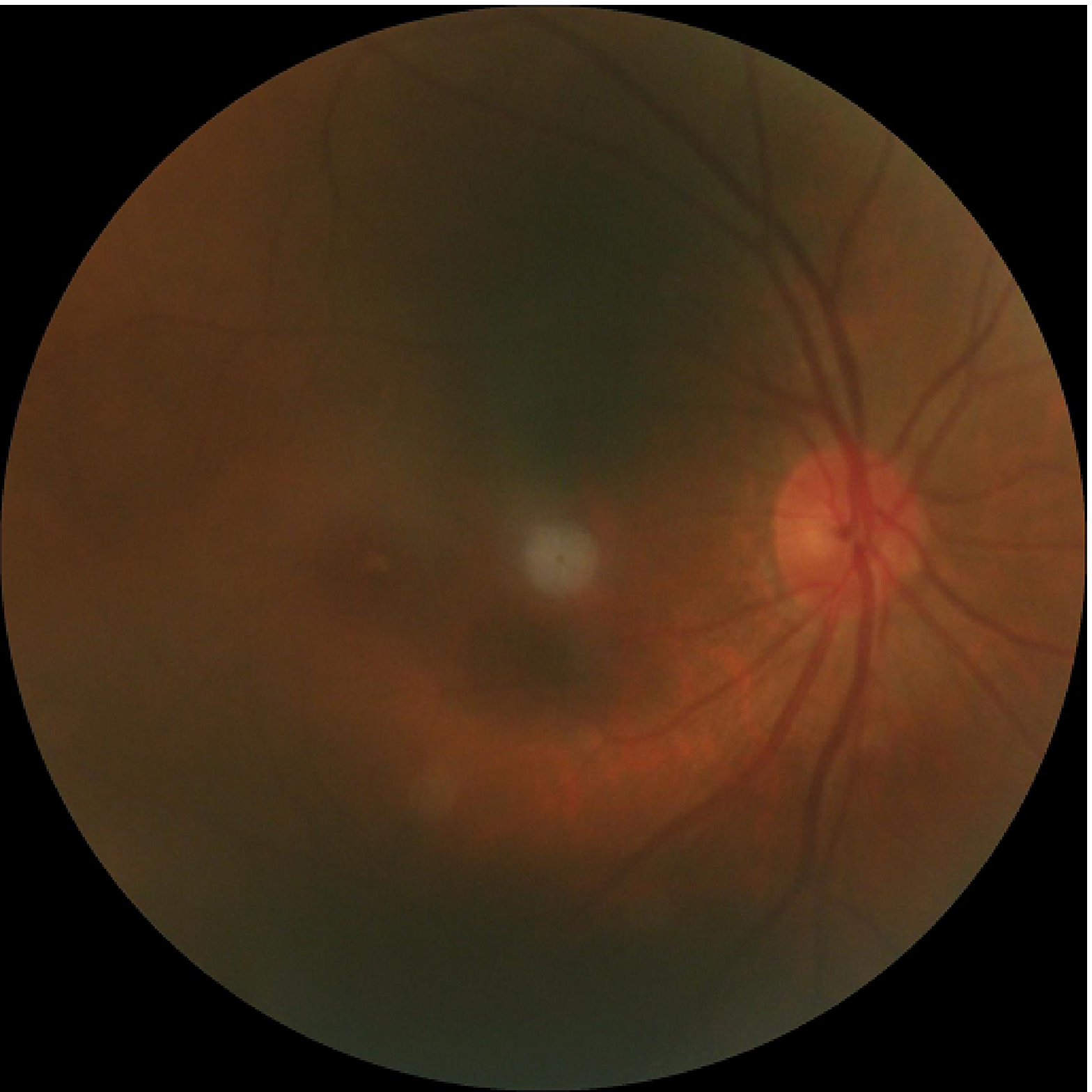}
\includegraphics[width=0.5in]{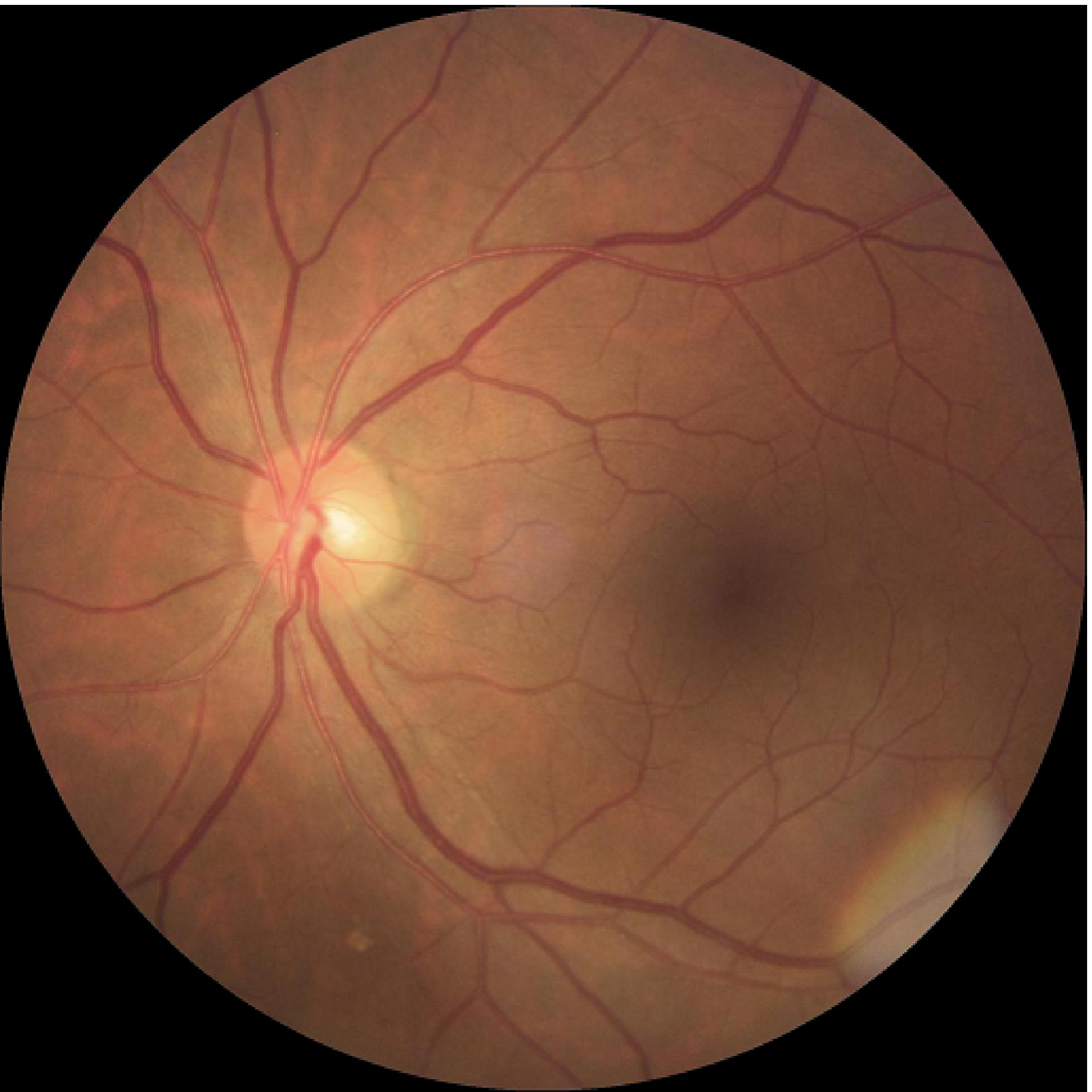}
\includegraphics[width=0.5in]{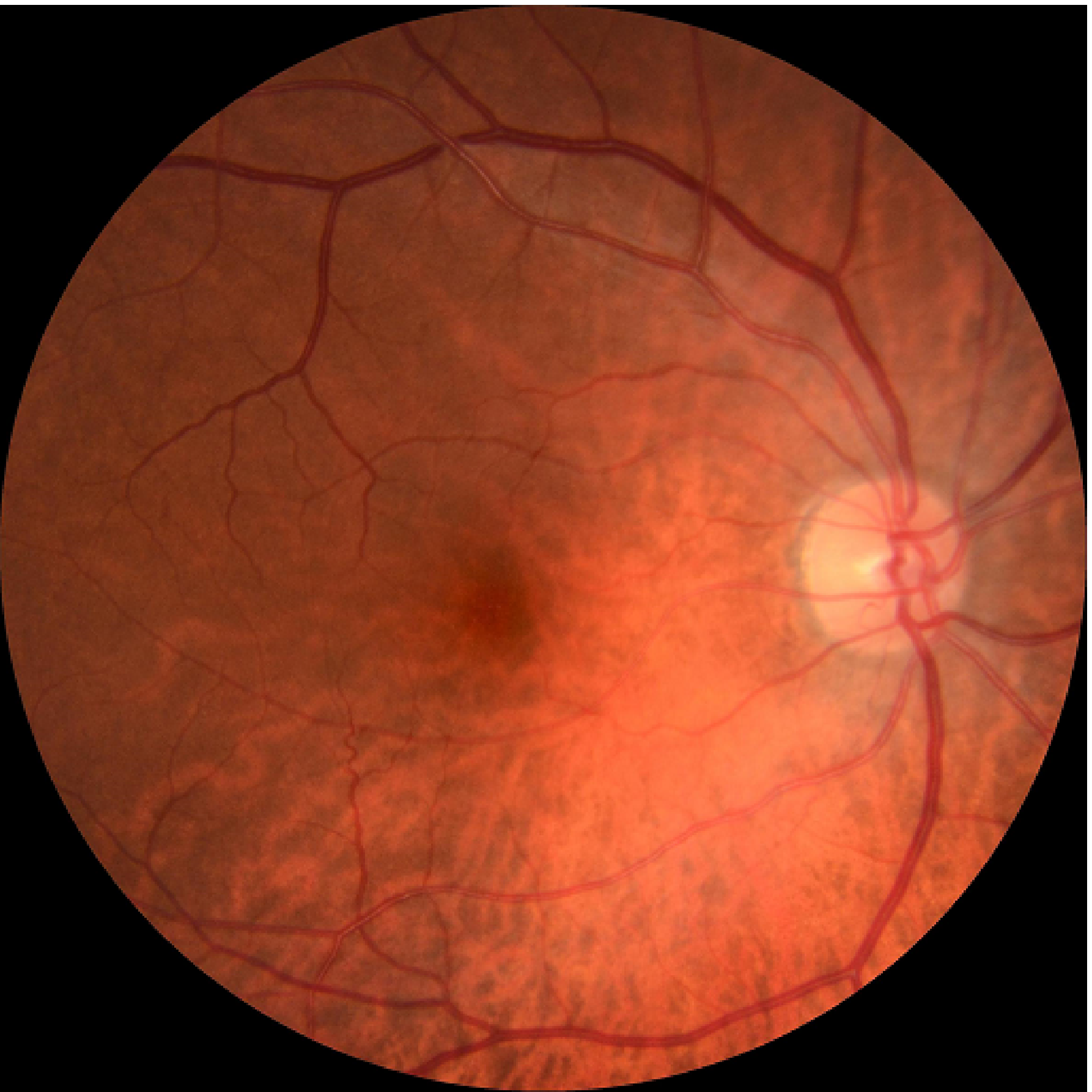}
\includegraphics[width=0.5in]{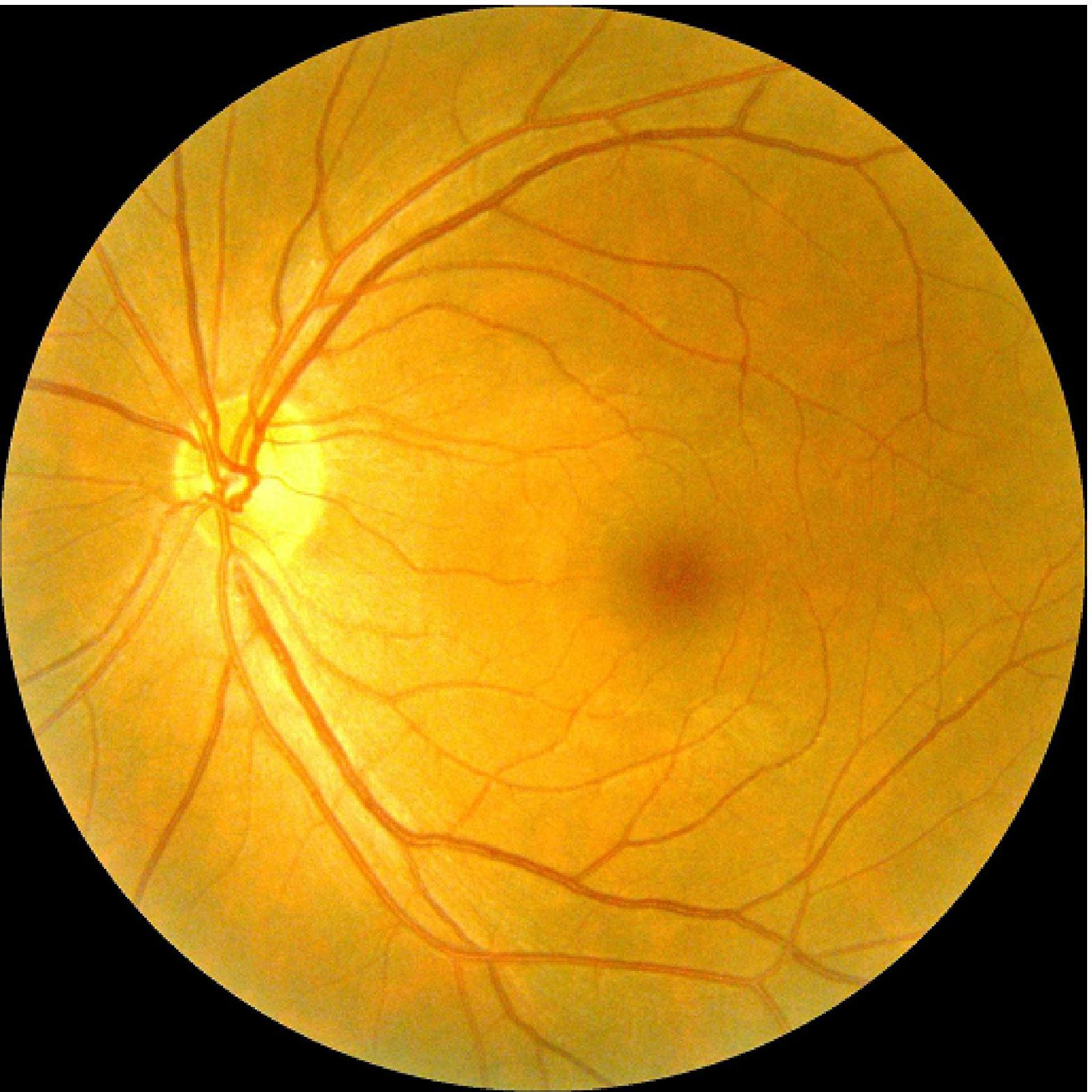}
\end{minipage}
}
\hspace{-0.25in}
\subfigure[D]{
\begin{minipage}[t]{0.125\linewidth}
\centering
\includegraphics[width=0.5in]{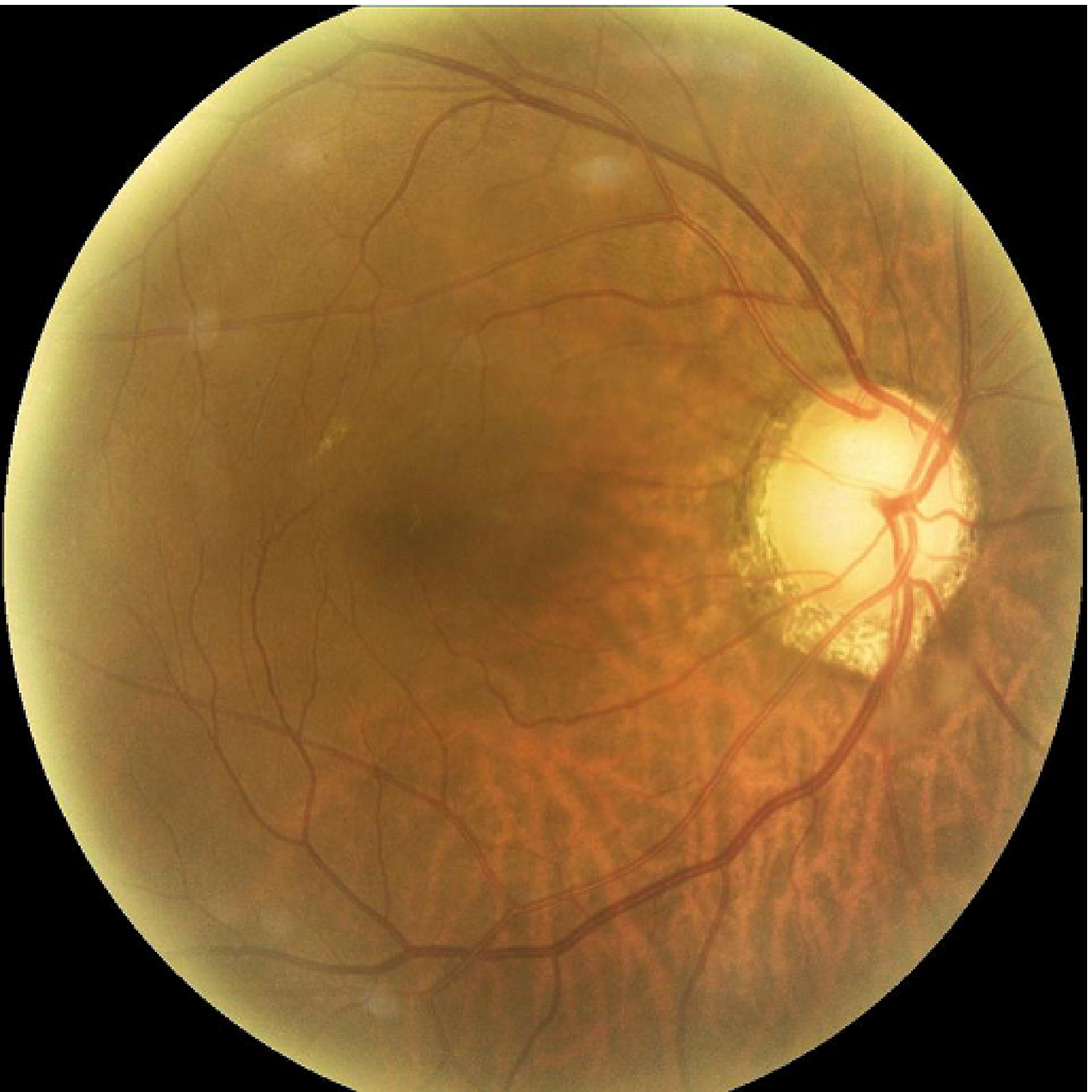}
\includegraphics[width=0.5in]{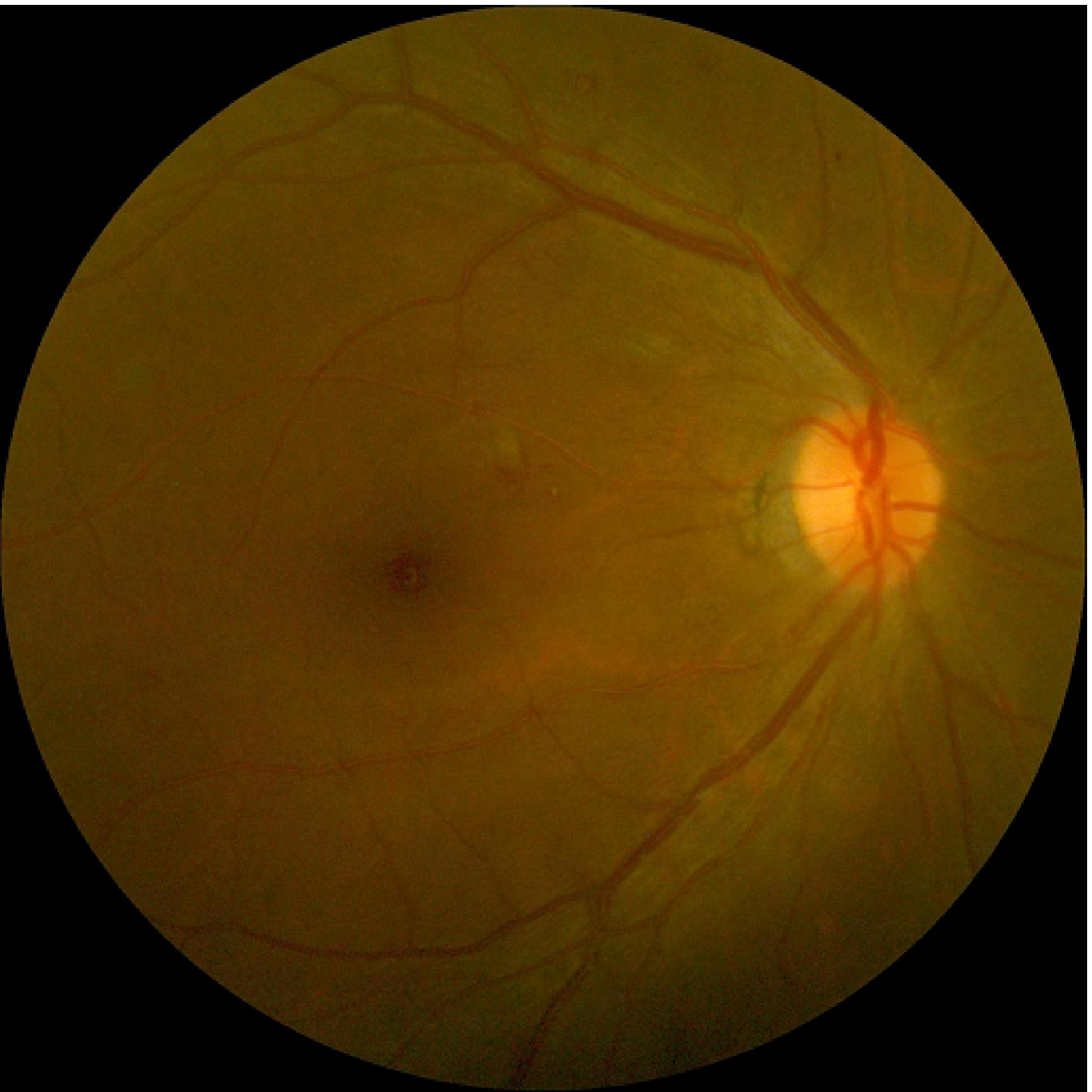}
\includegraphics[width=0.5in]{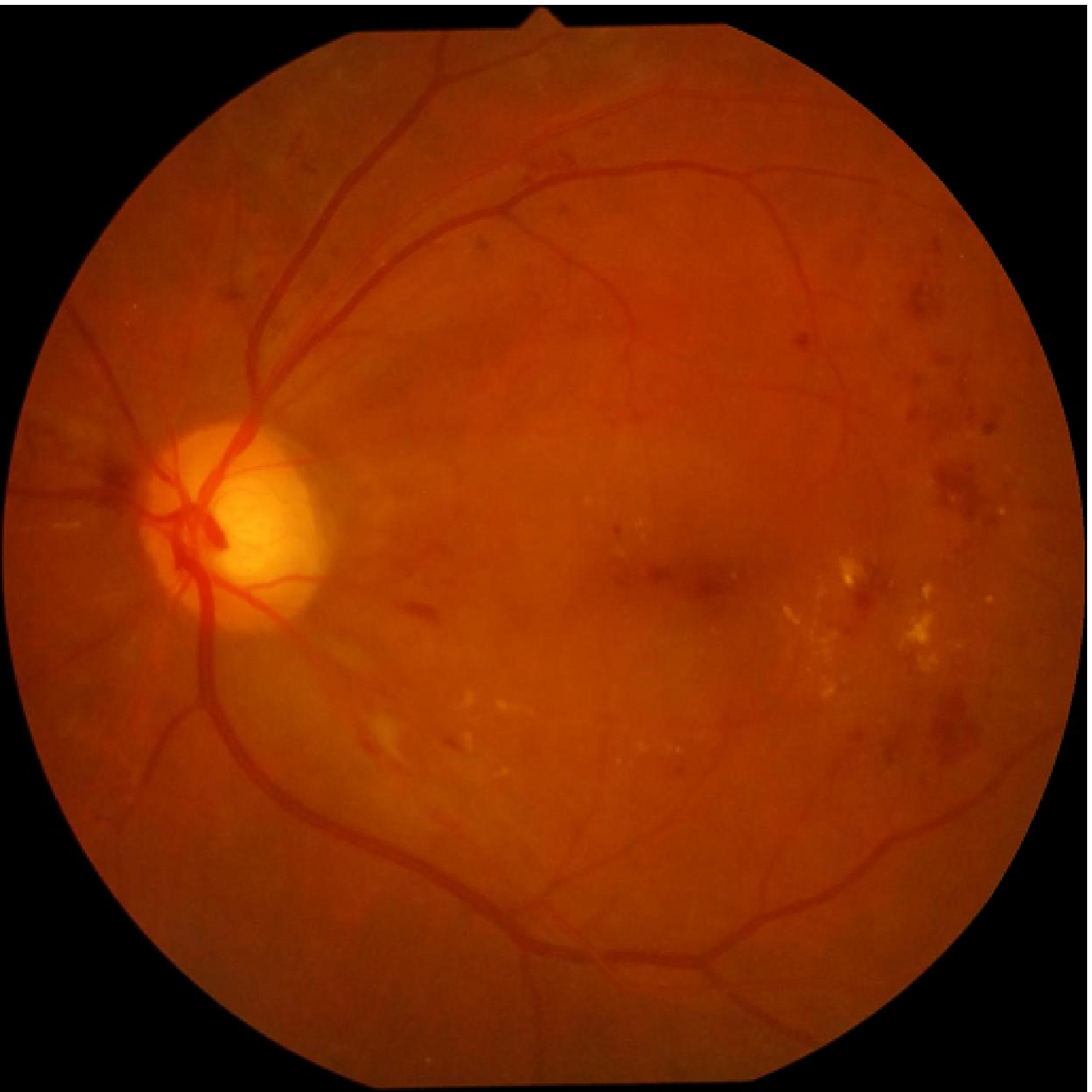}
\includegraphics[width=0.5in]{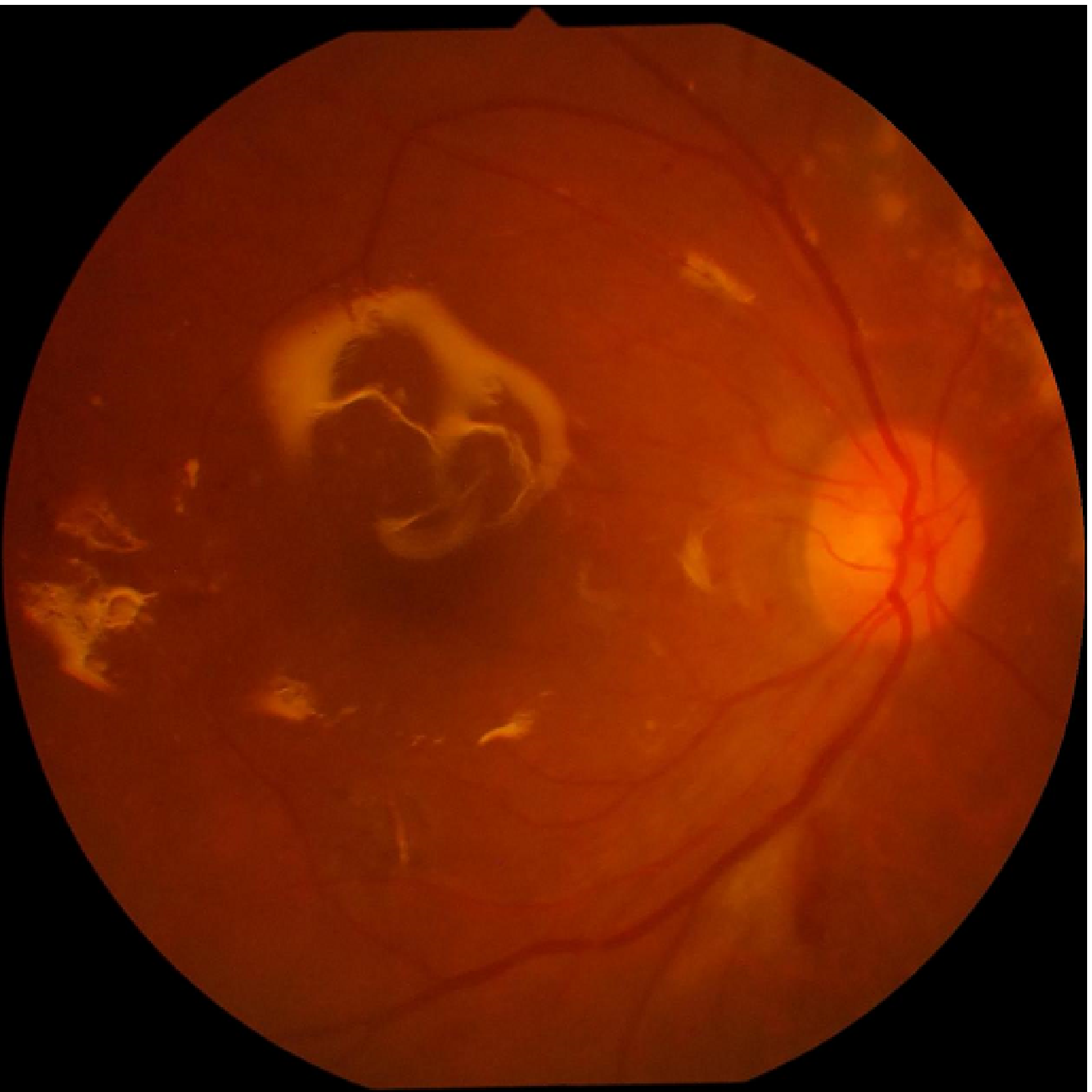}
\end{minipage}
}
\hspace{-0.25in}
\subfigure[G]{
\begin{minipage}[t]{0.125\linewidth}
\centering
\includegraphics[width=0.5in]{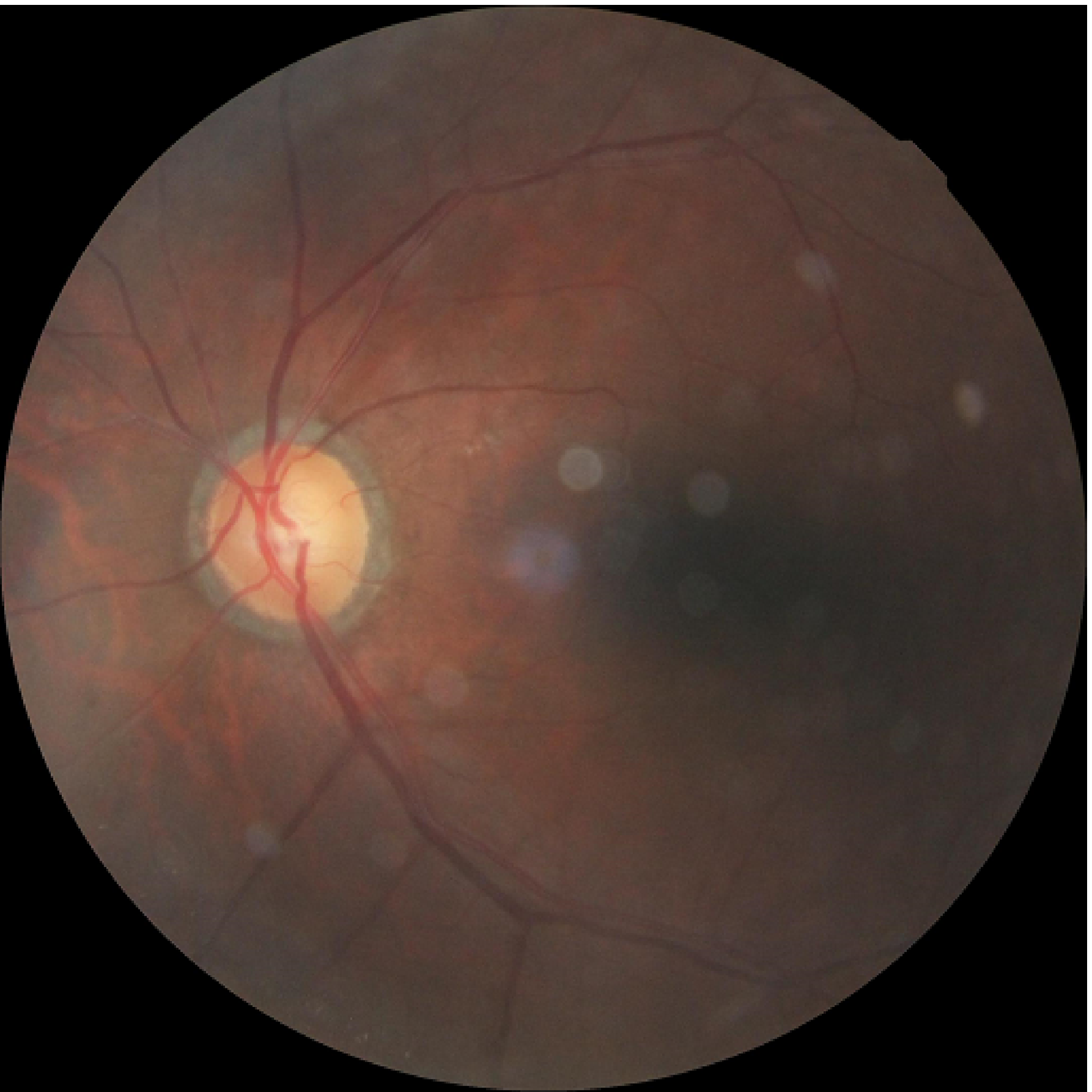}
\includegraphics[width=0.5in]{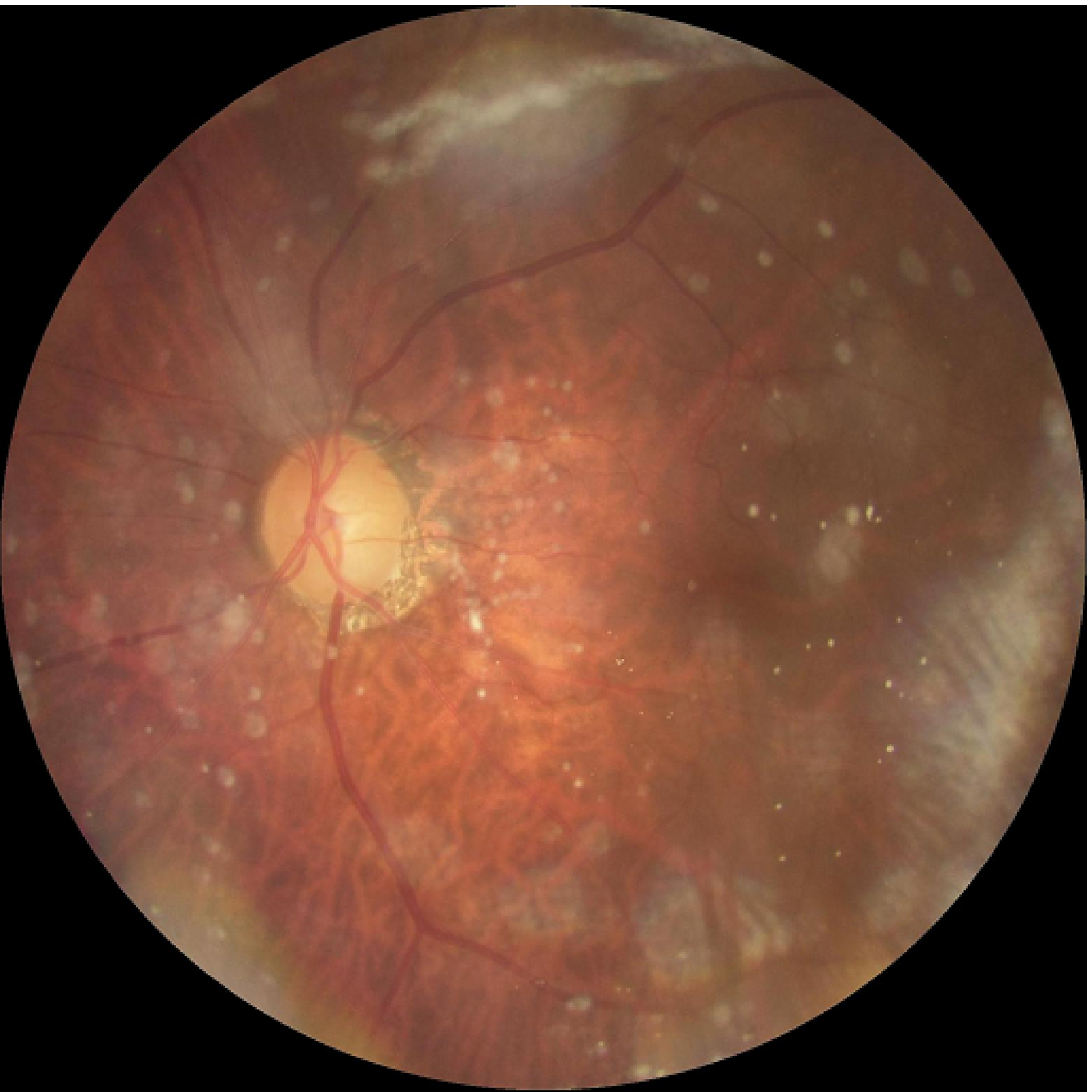}
\includegraphics[width=0.5in]{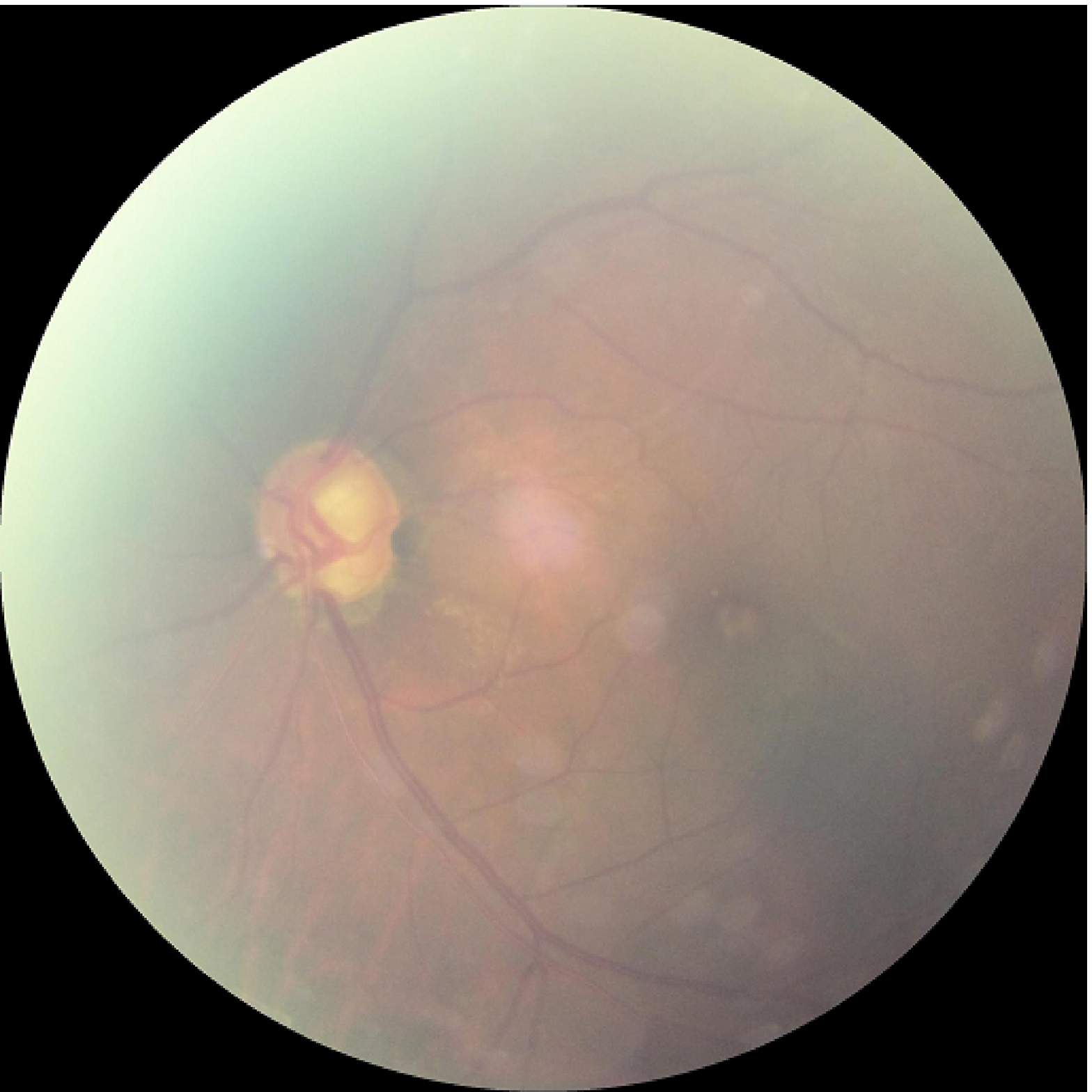}
\includegraphics[width=0.5in]{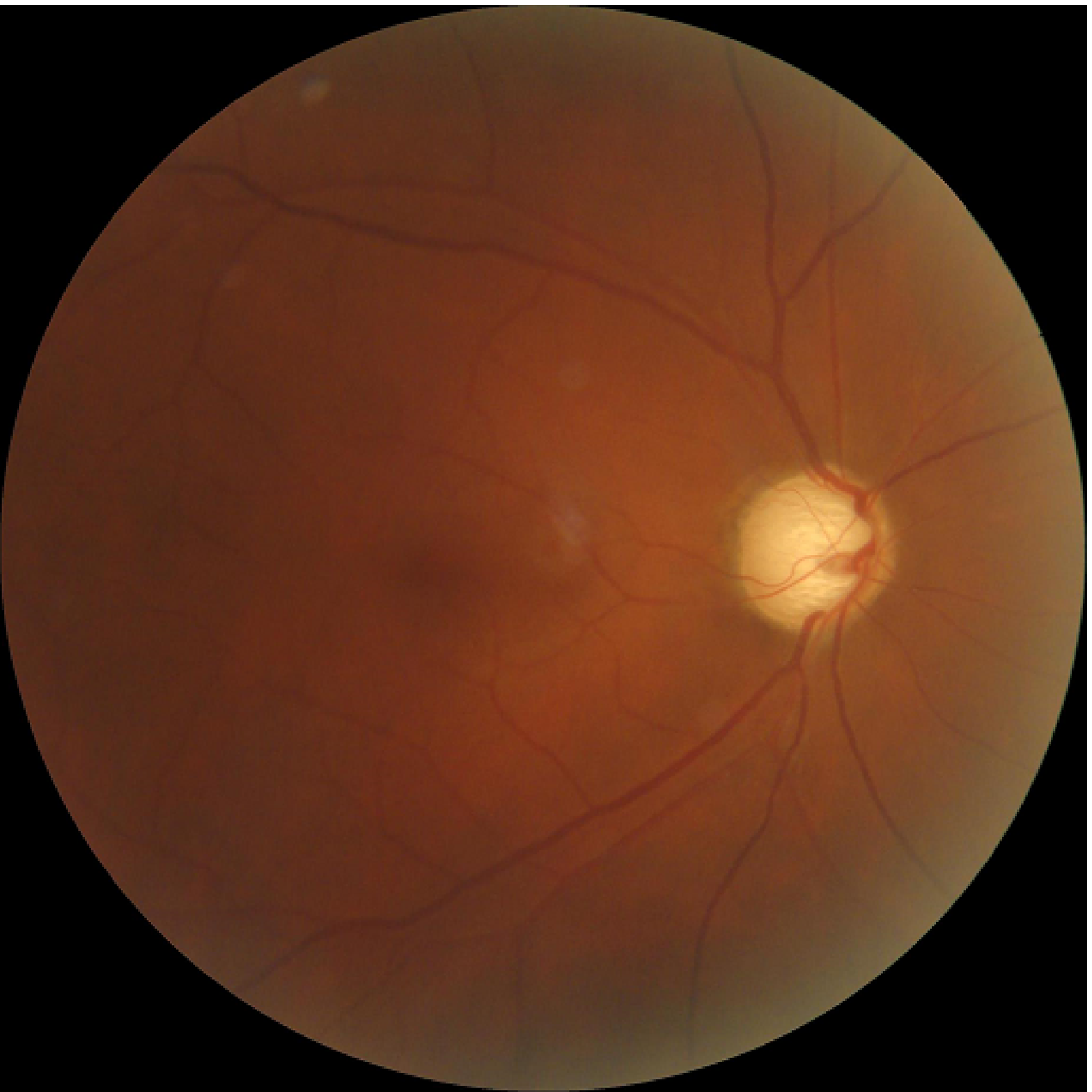}
\end{minipage}
}
\hspace{-0.25in}
\subfigure[C]{
\begin{minipage}[t]{0.125\linewidth}
\centering
\includegraphics[width=0.5in]{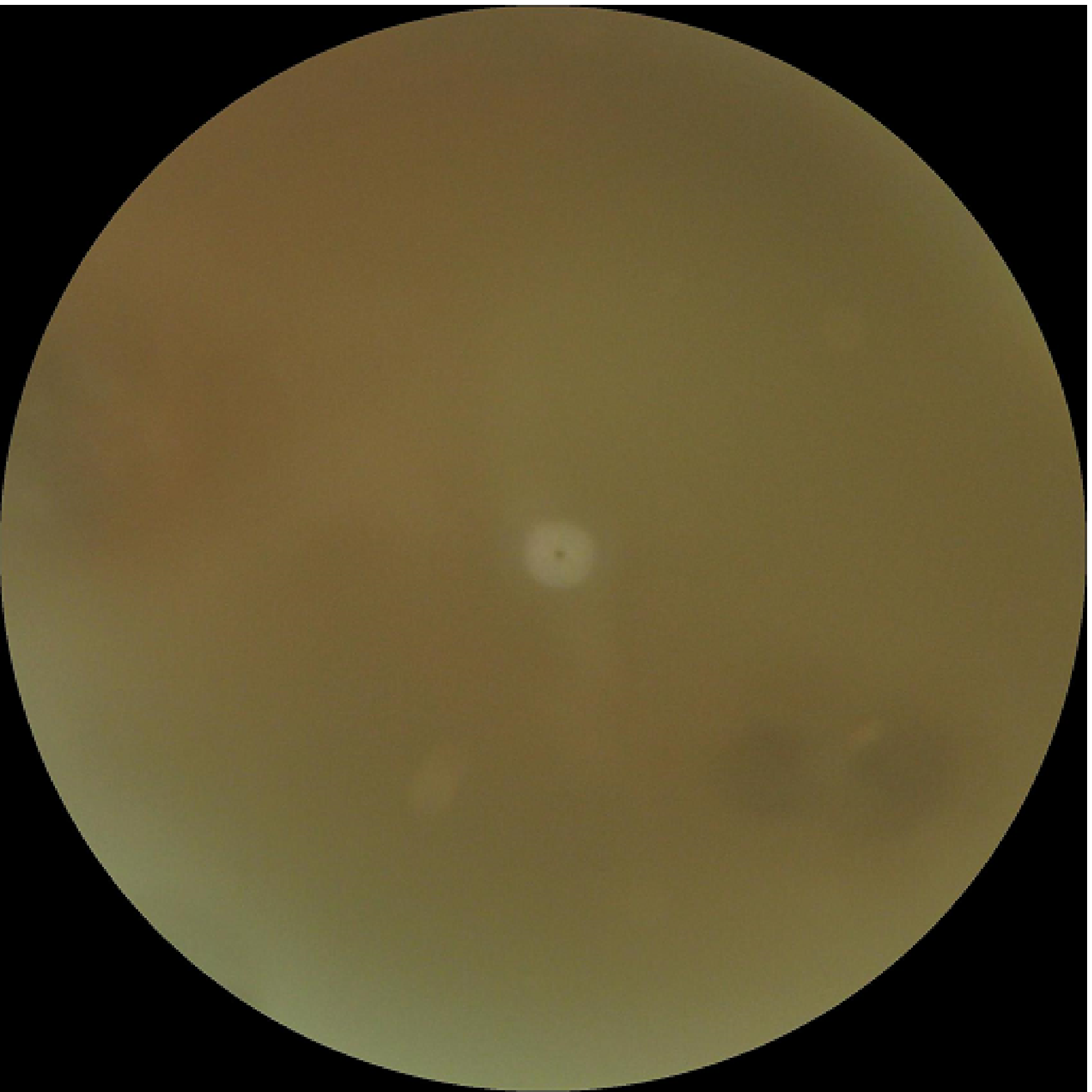}
\includegraphics[width=0.5in]{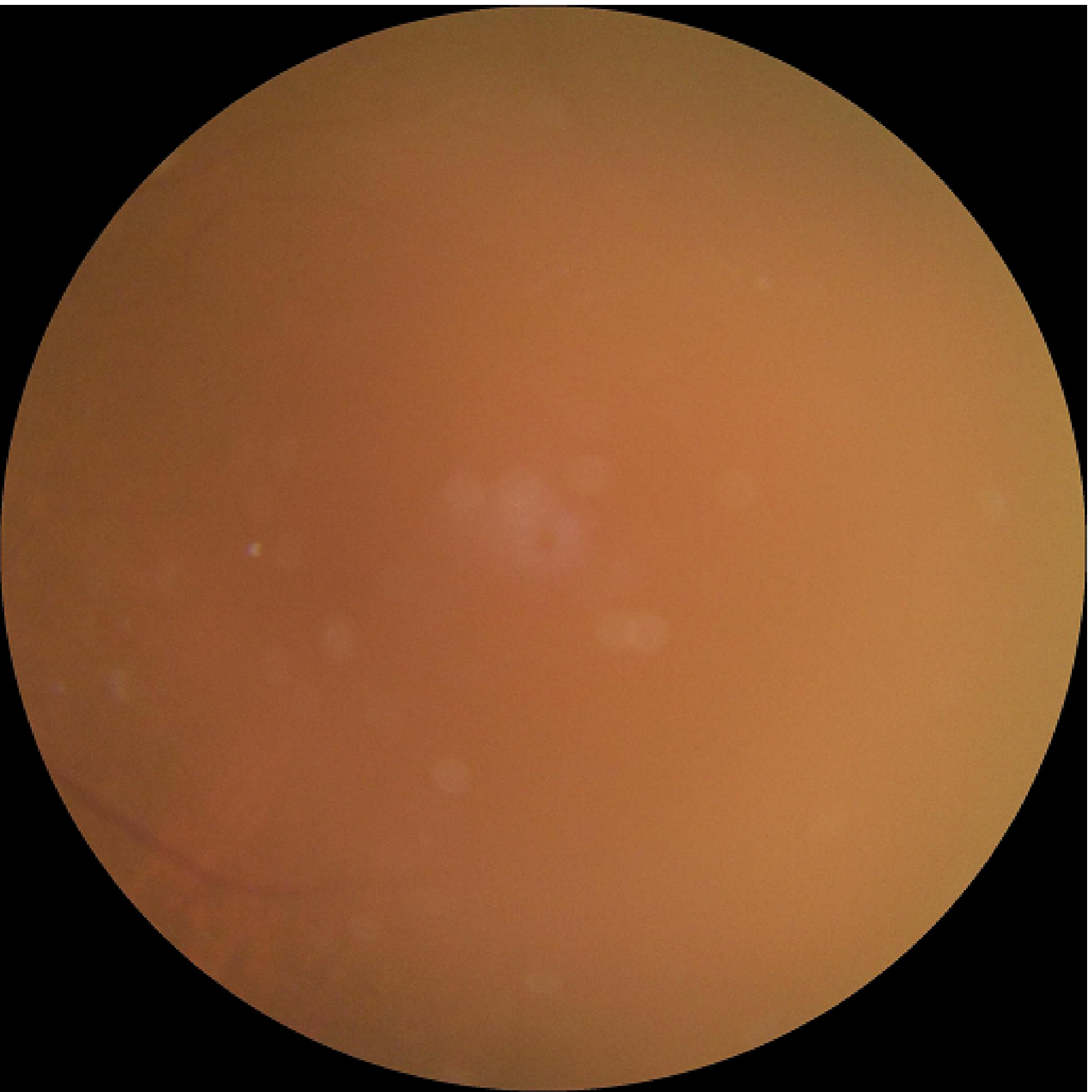}
\includegraphics[width=0.5in]{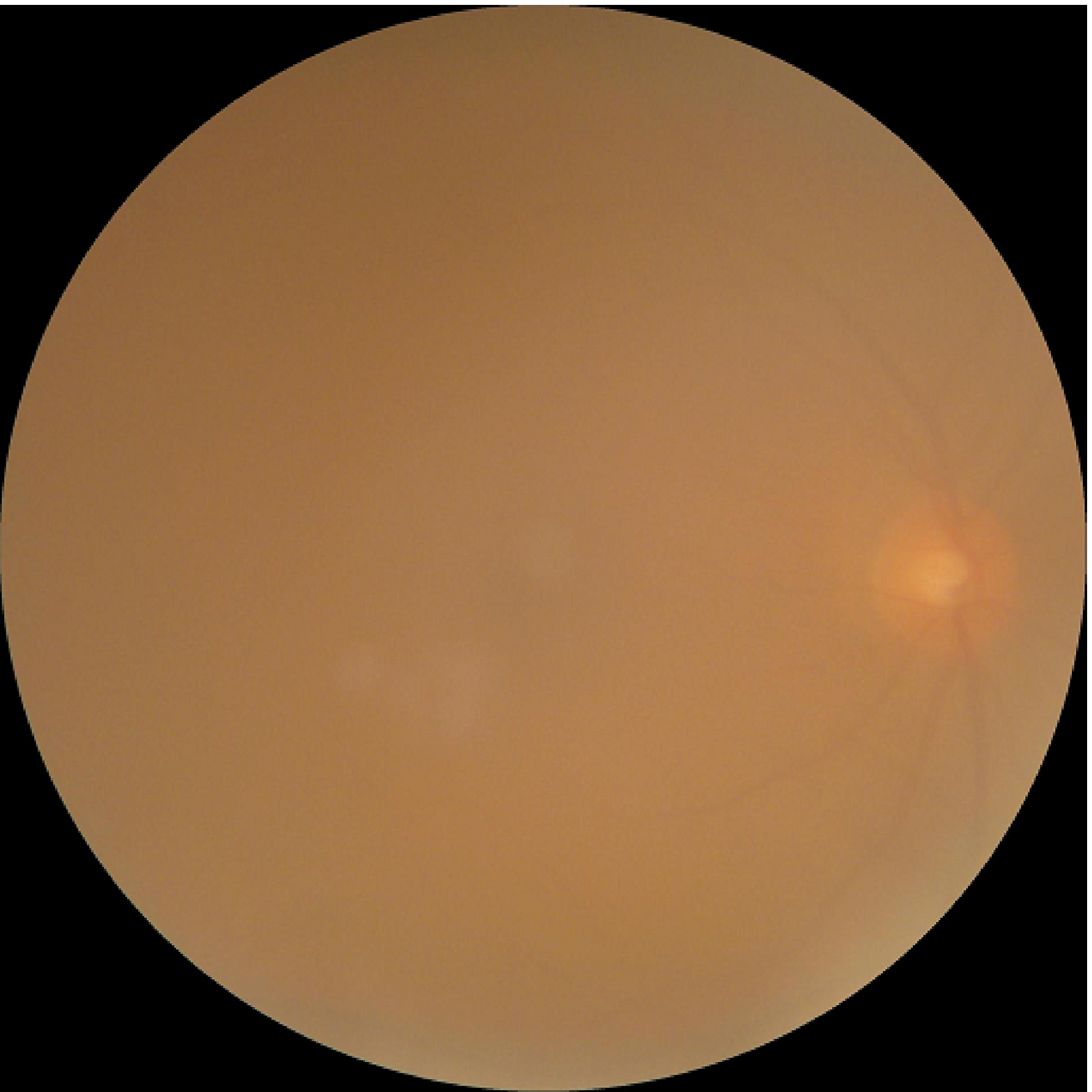}
\includegraphics[width=0.5in]{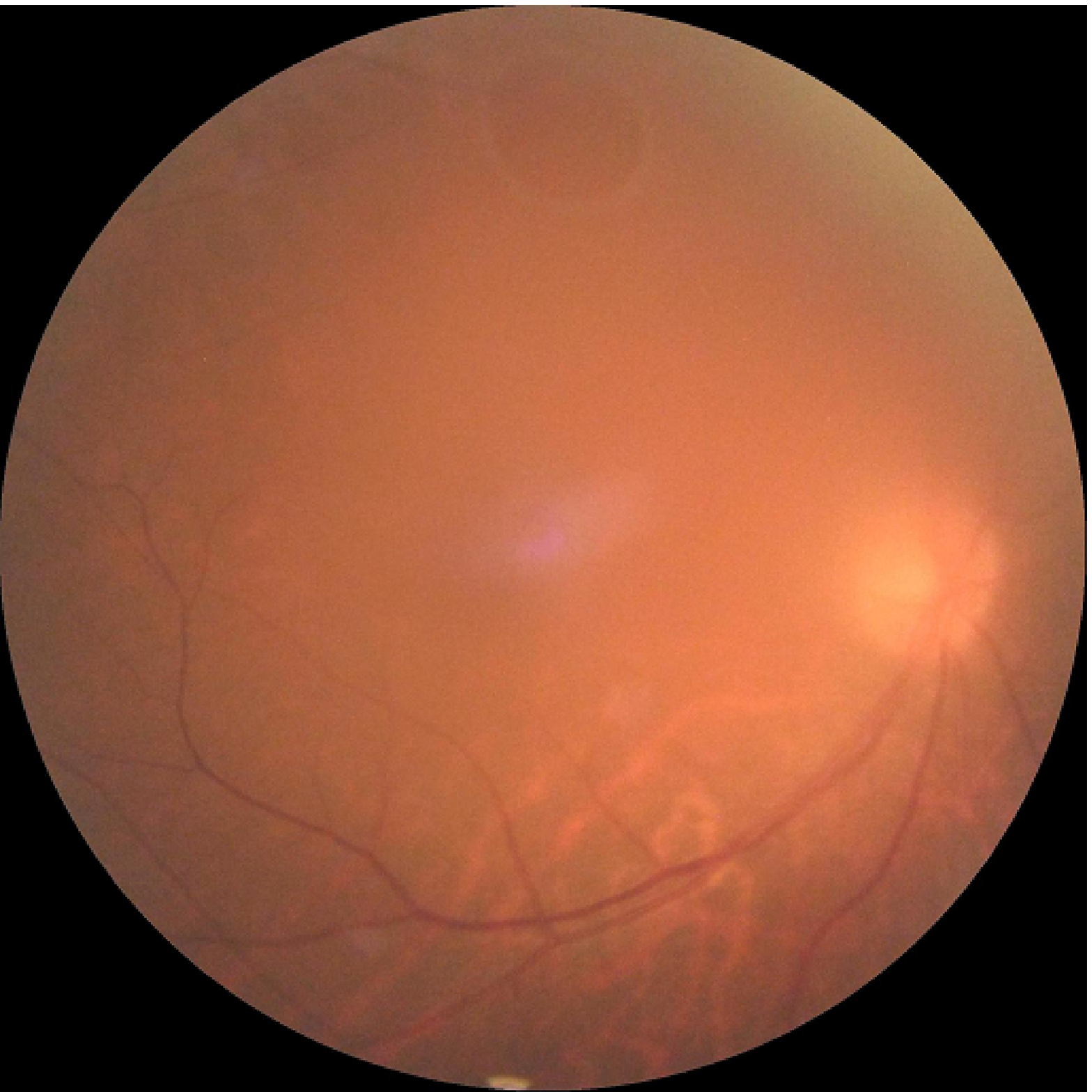}
\end{minipage}
}
\hspace{-0.25in}
\subfigure[A]{
\begin{minipage}[t]{0.125\linewidth}
\centering
\includegraphics[width=0.5in]{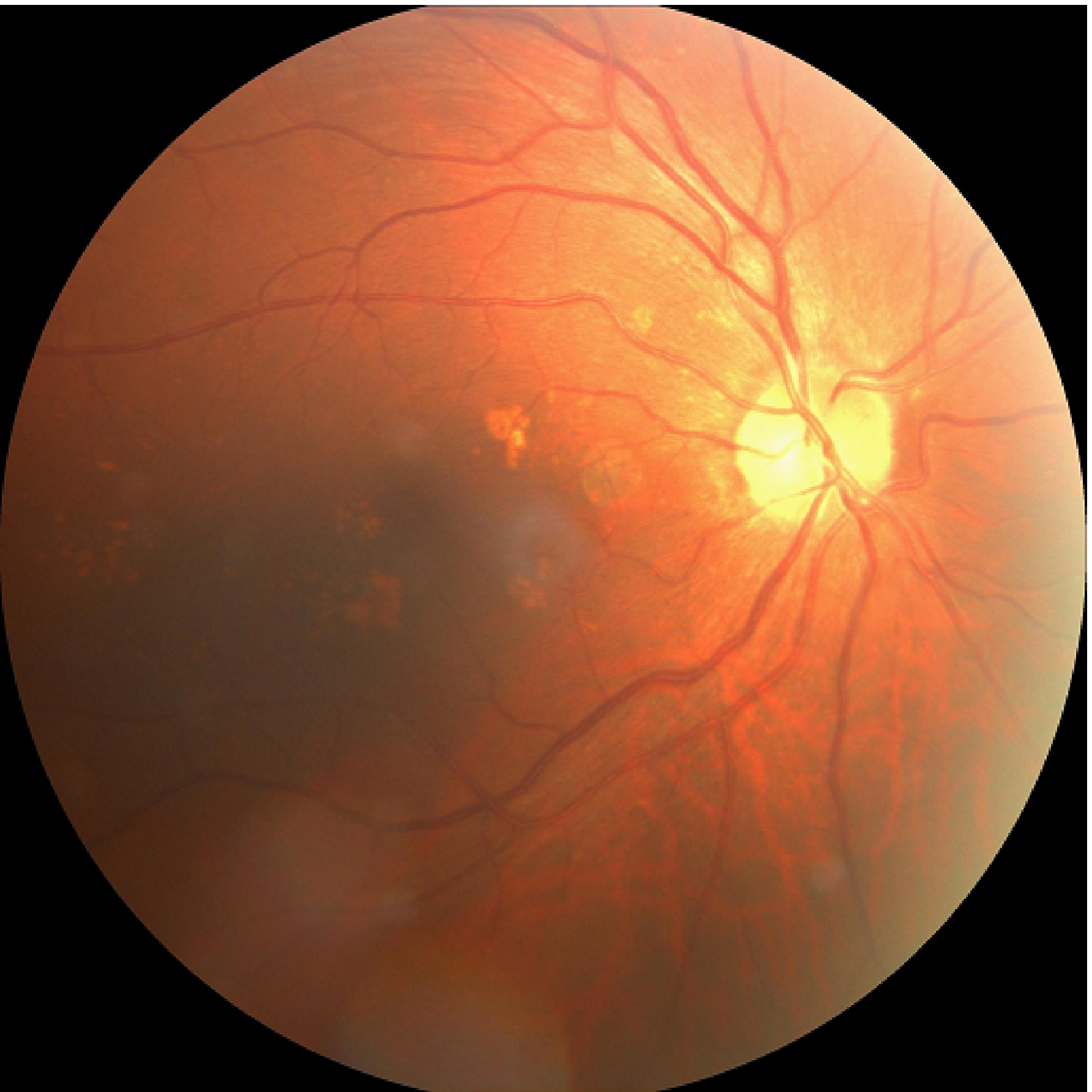}
\includegraphics[width=0.5in]{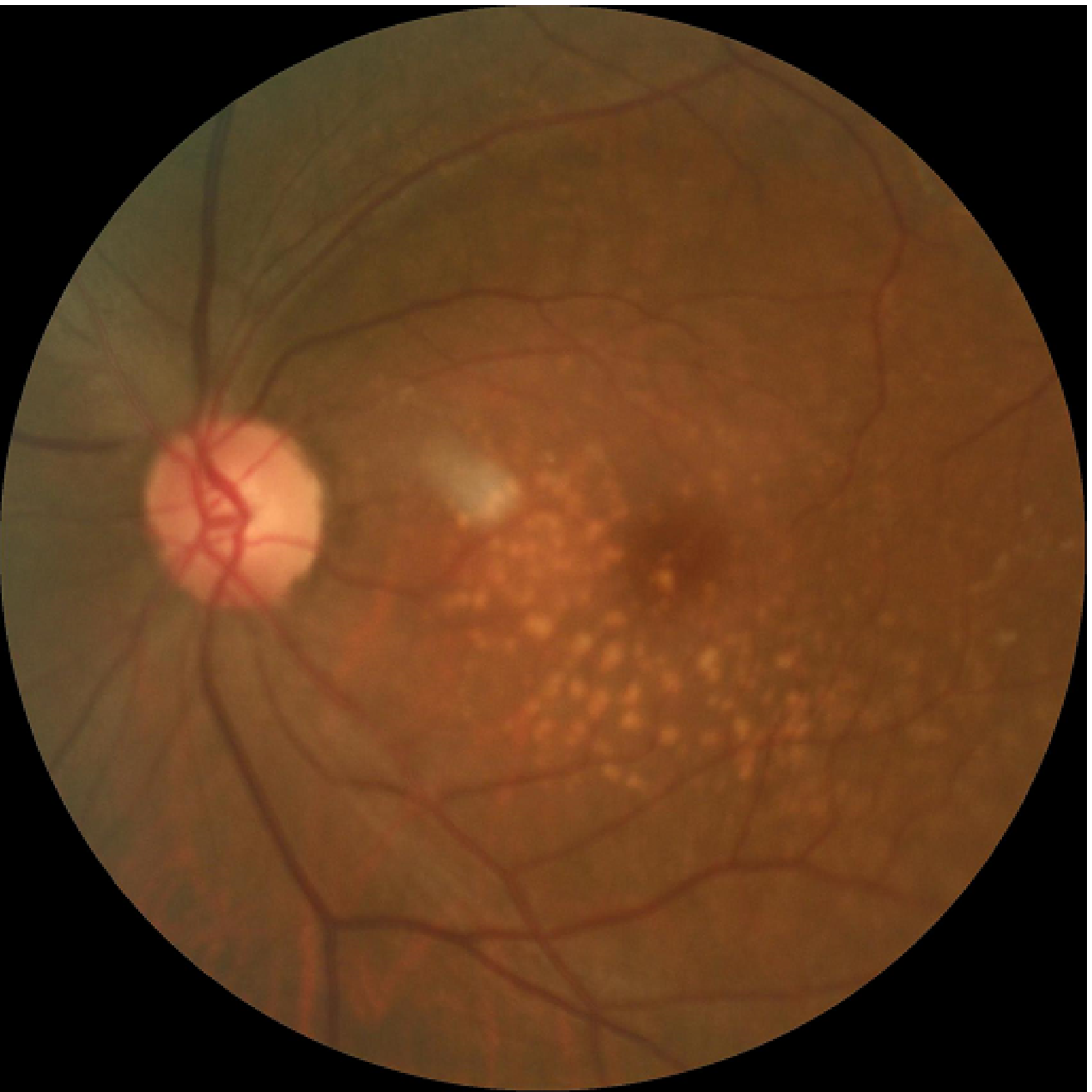}
\includegraphics[width=0.5in]{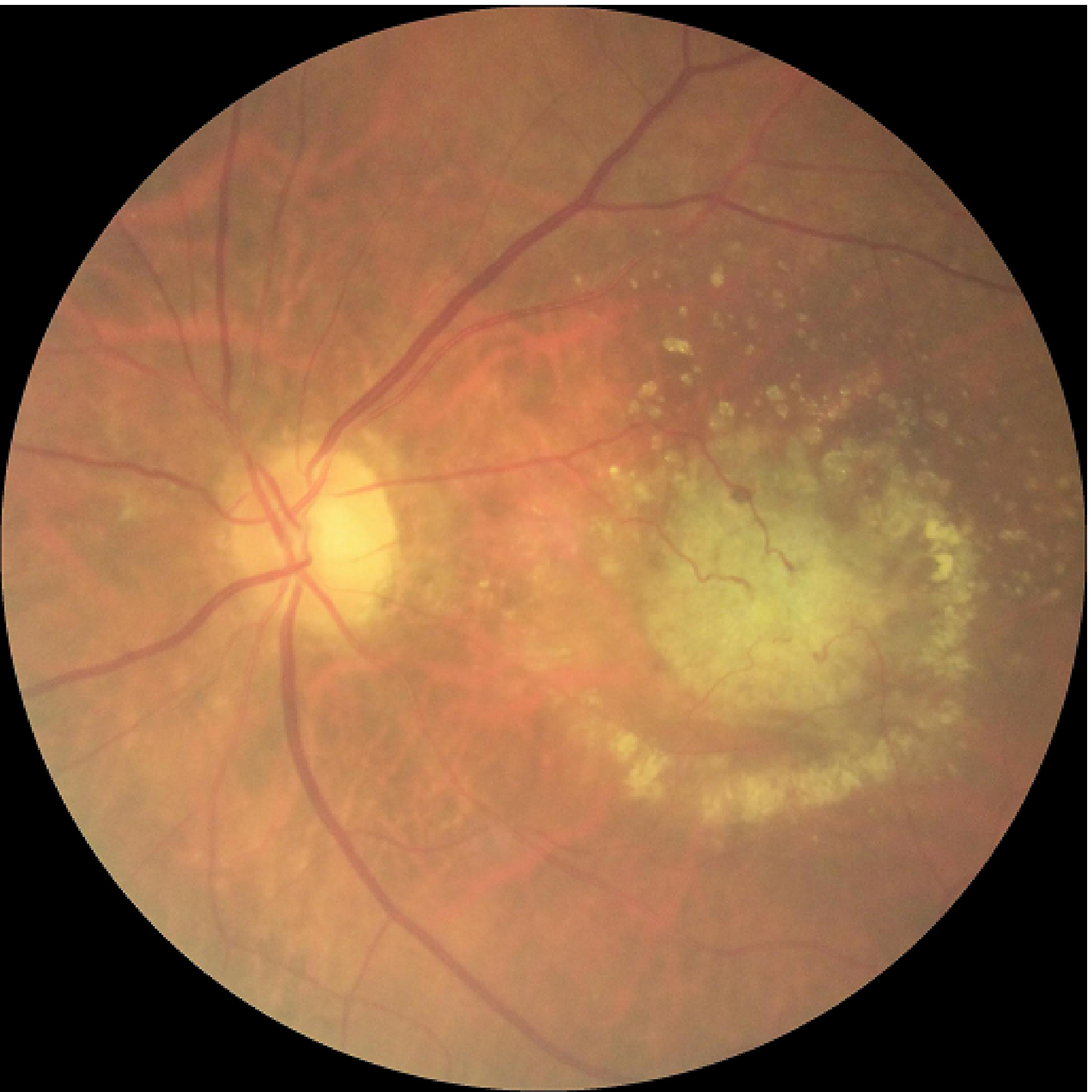}
\includegraphics[width=0.5in]{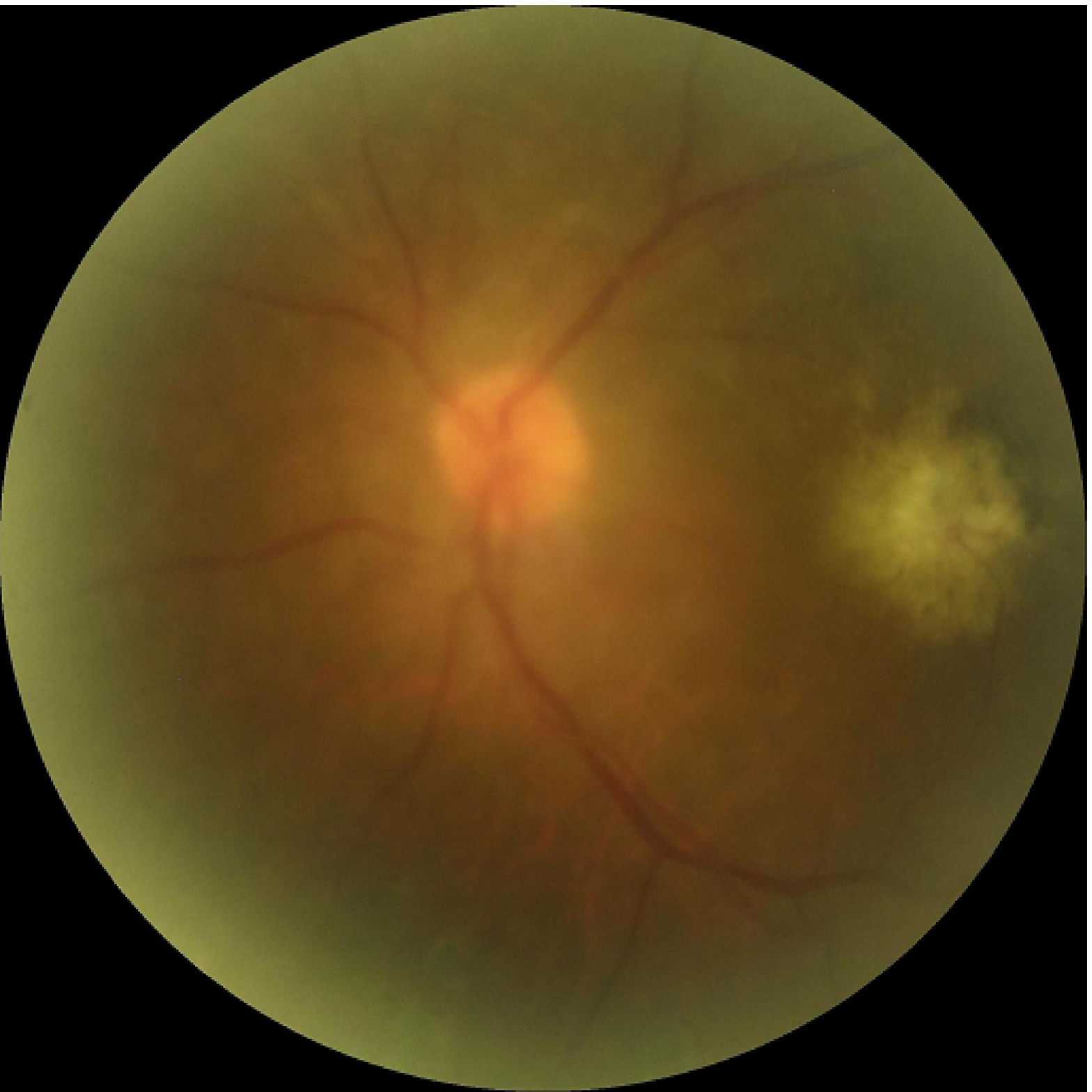}
\end{minipage}
}
\hspace{-0.25in}
\subfigure[H]{
\begin{minipage}[t]{0.125\linewidth}
\centering
\includegraphics[width=0.5in]{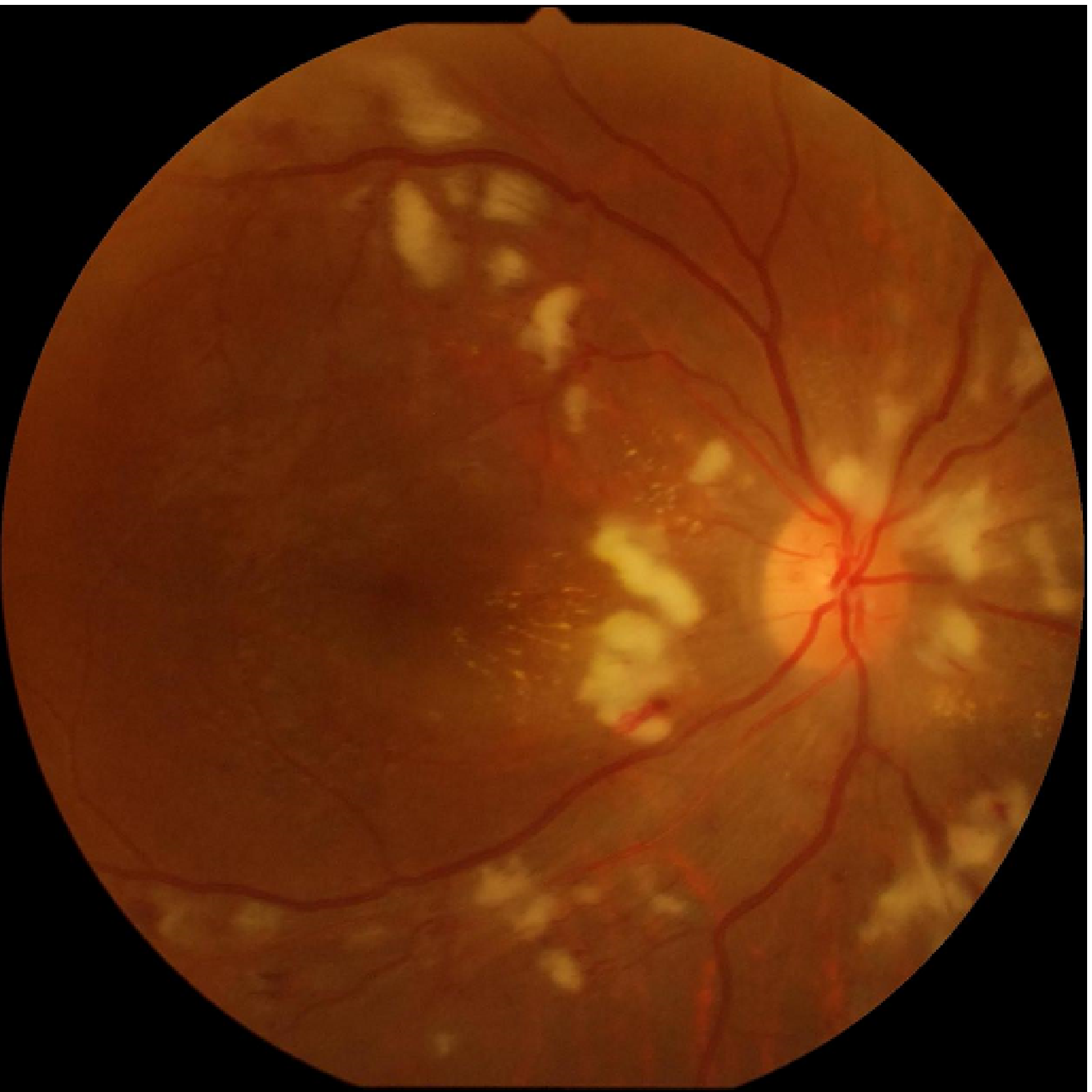}
\includegraphics[width=0.5in]{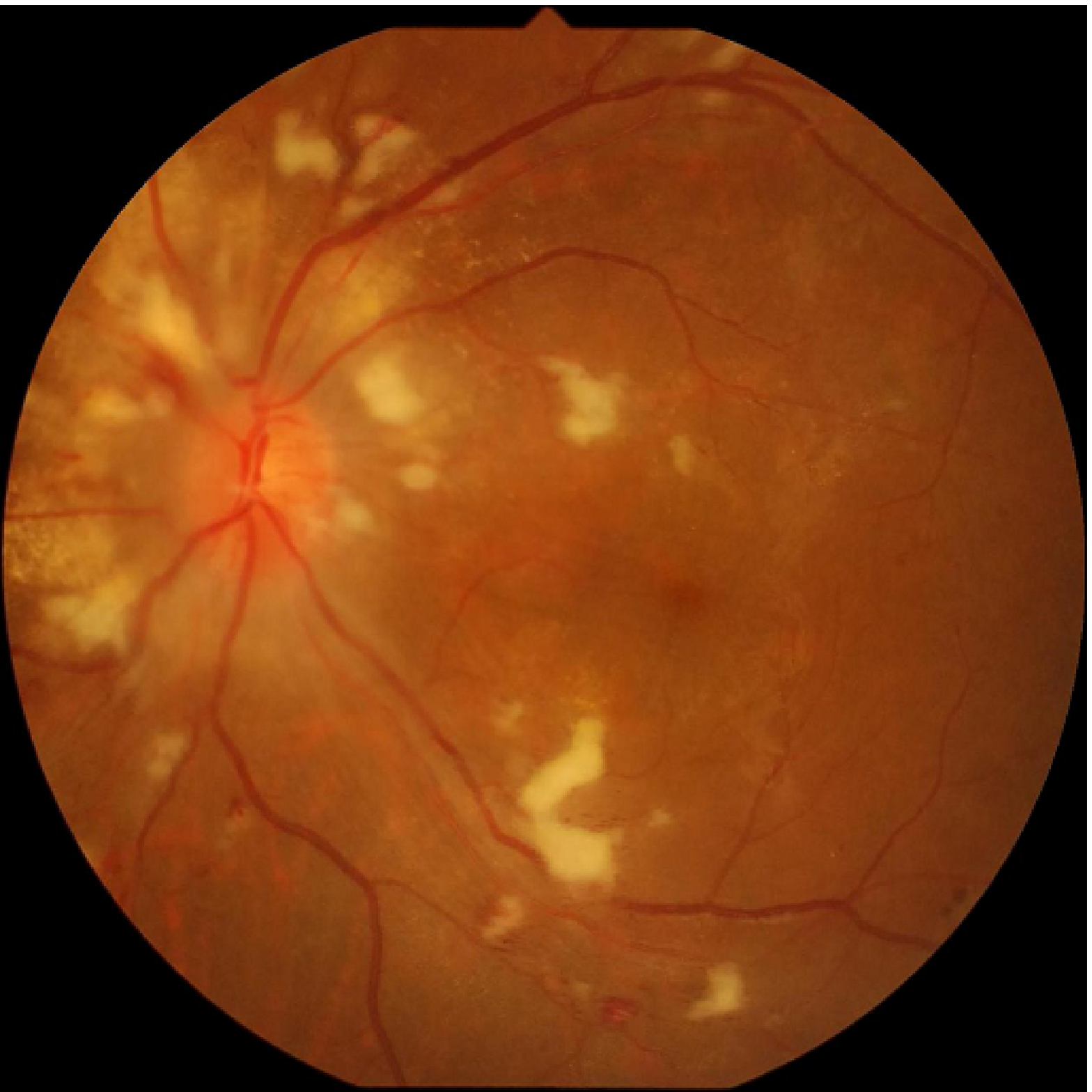}
\includegraphics[width=0.5in]{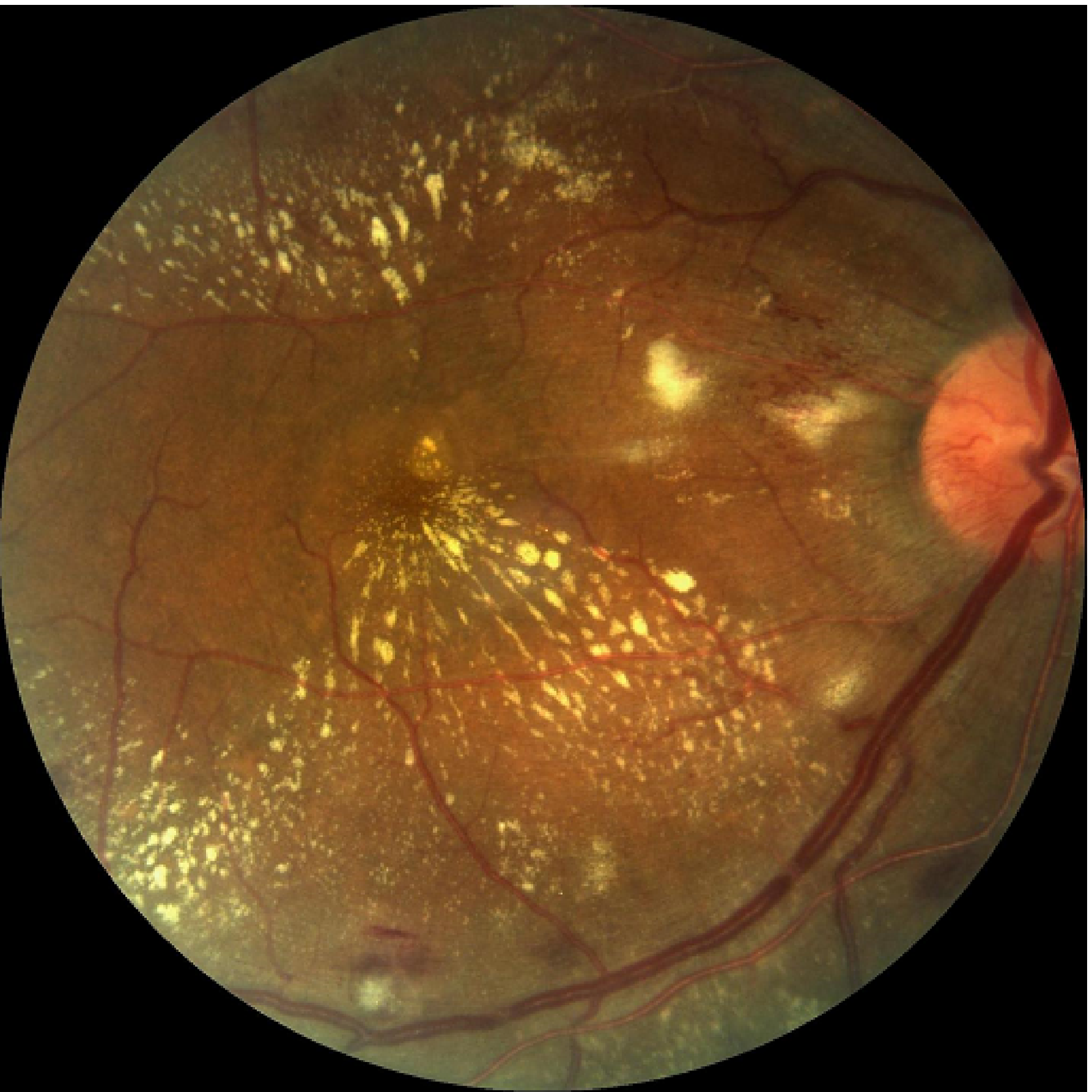}
\includegraphics[width=0.5in]{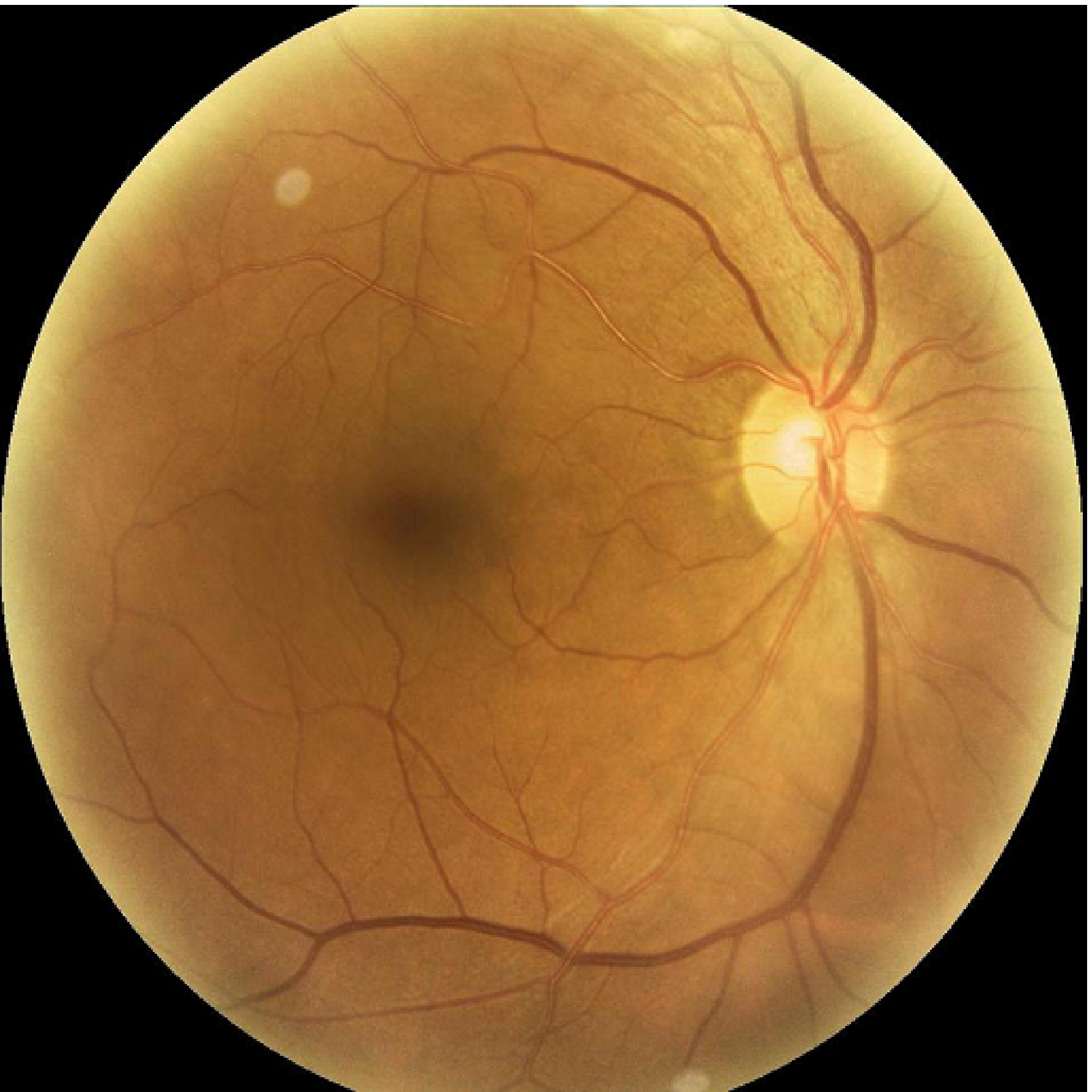}
\end{minipage}
}
\hspace{-0.25in}
\subfigure[M]{
\begin{minipage}[t]{0.125\linewidth}
\centering
\includegraphics[width=0.5in]{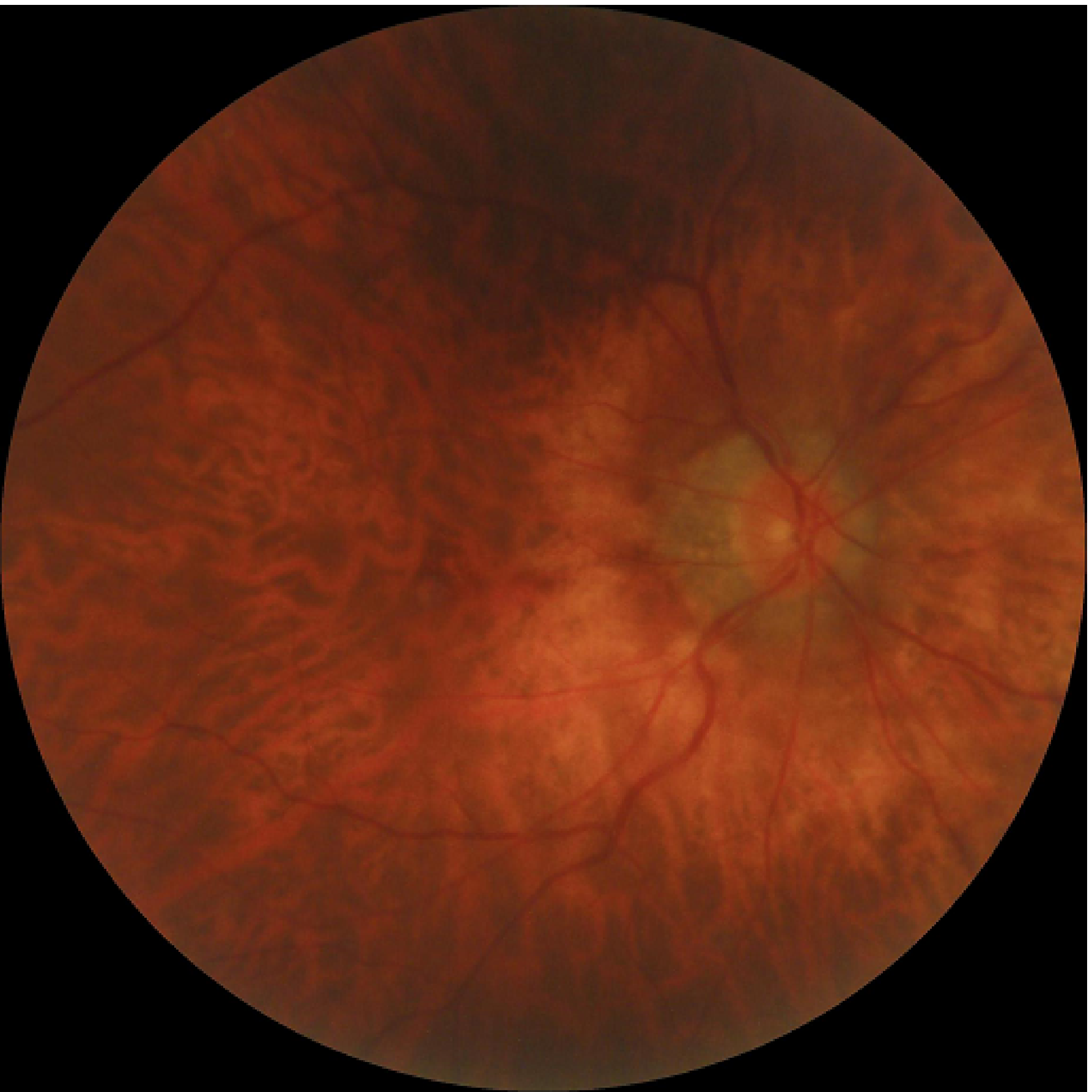}
\includegraphics[width=0.5in]{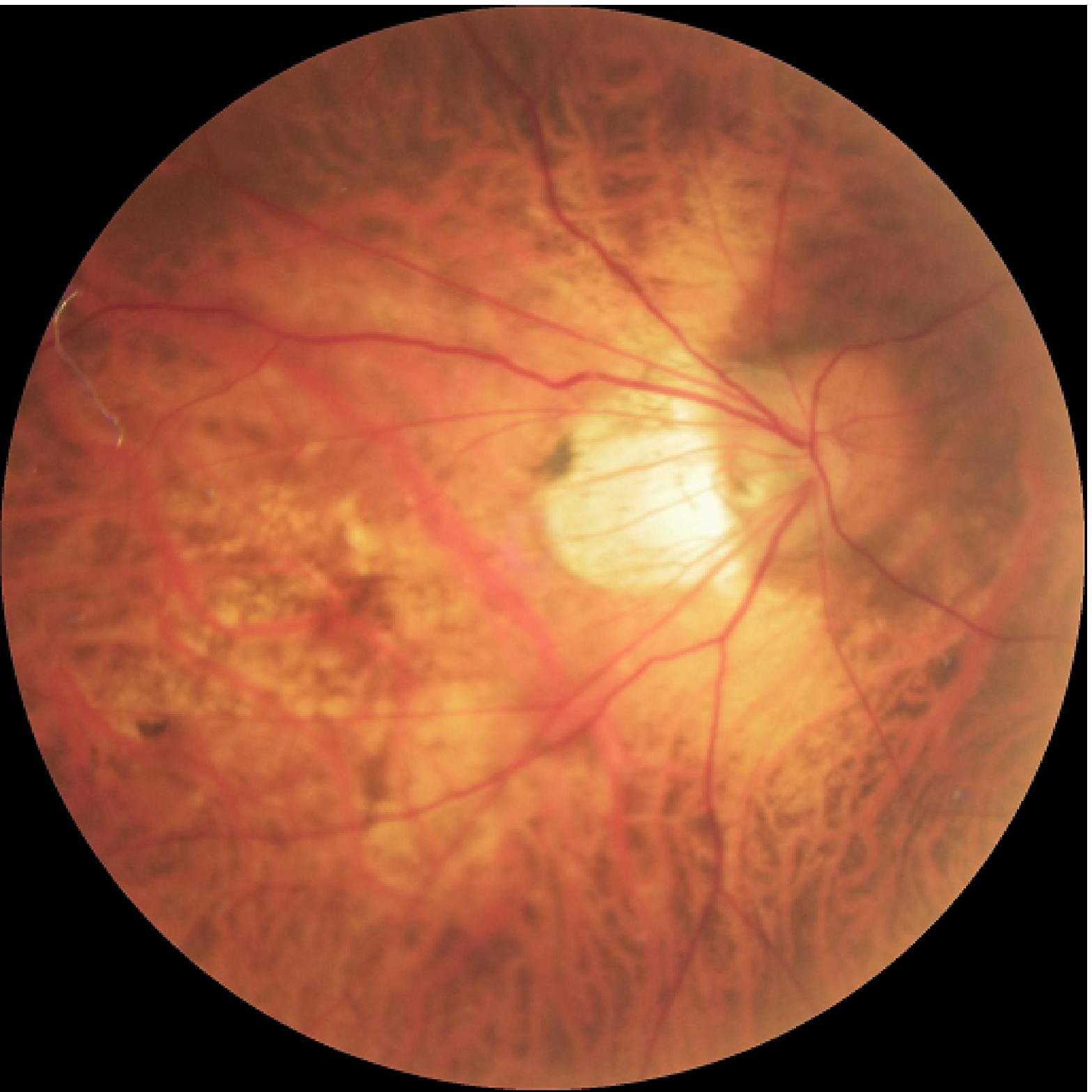}
\includegraphics[width=0.5in]{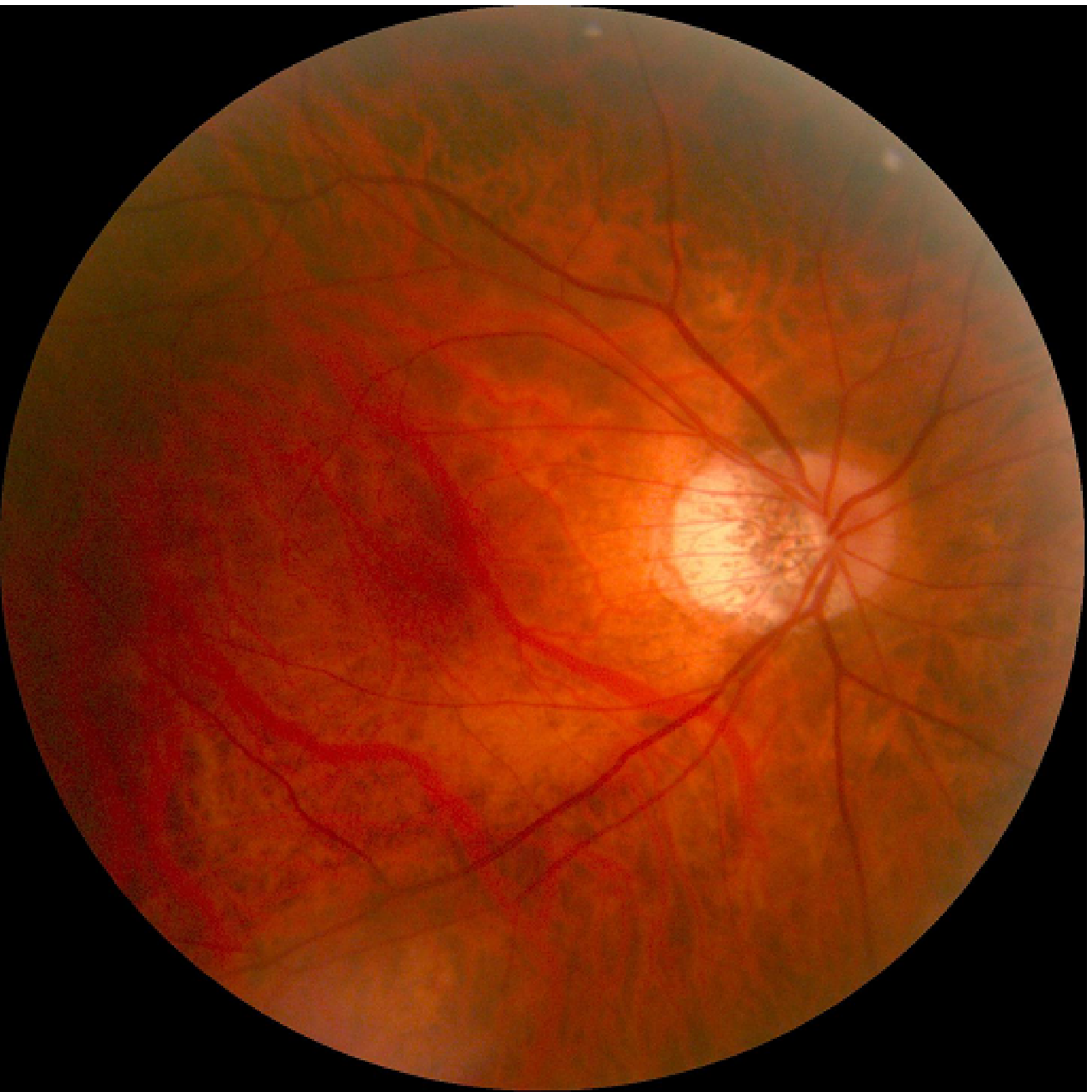}
\includegraphics[width=0.5in]{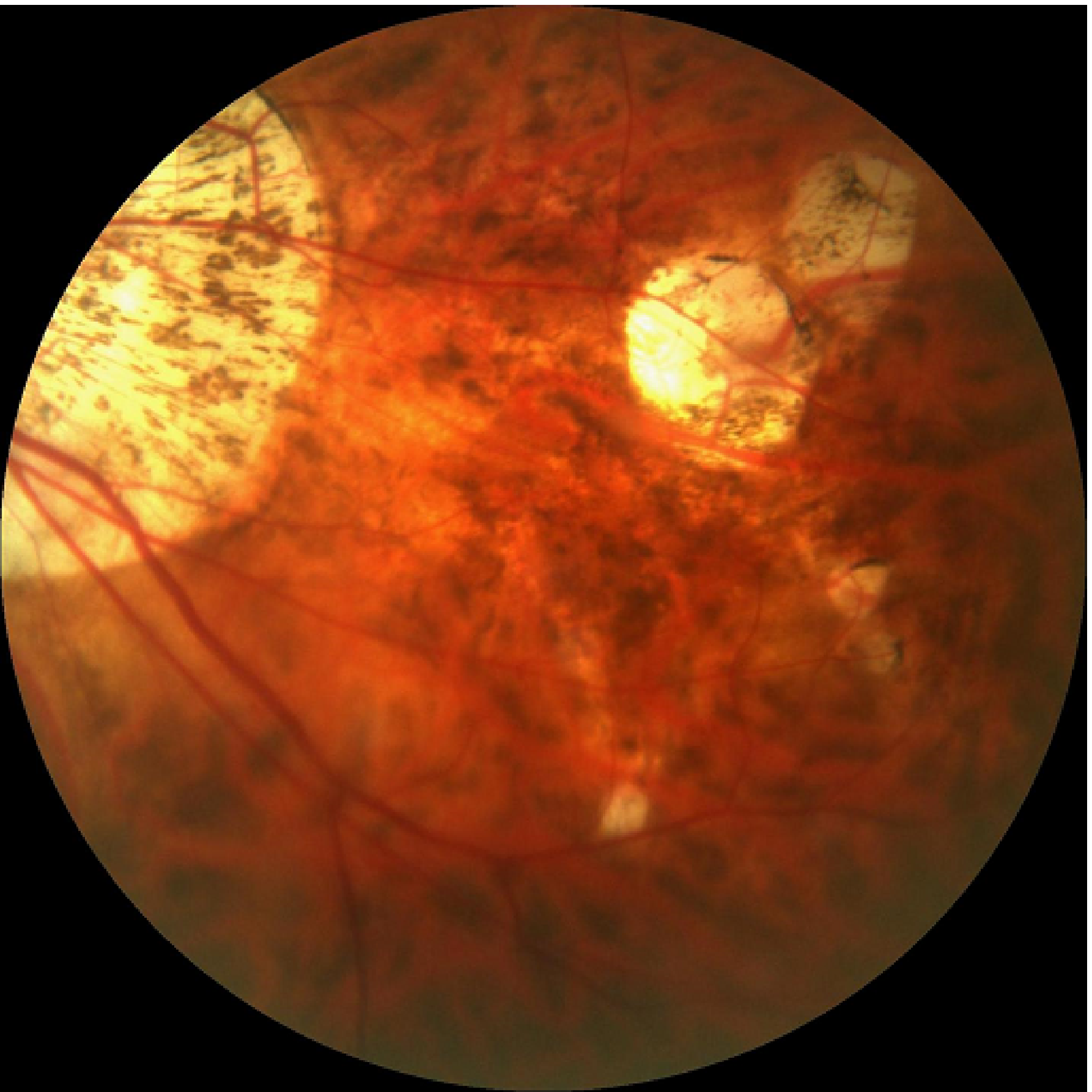}
\end{minipage}
}
\hspace{-0.25in}
\subfigure[O]{
\begin{minipage}[t]{0.125\linewidth}
\centering
\includegraphics[width=0.5in]{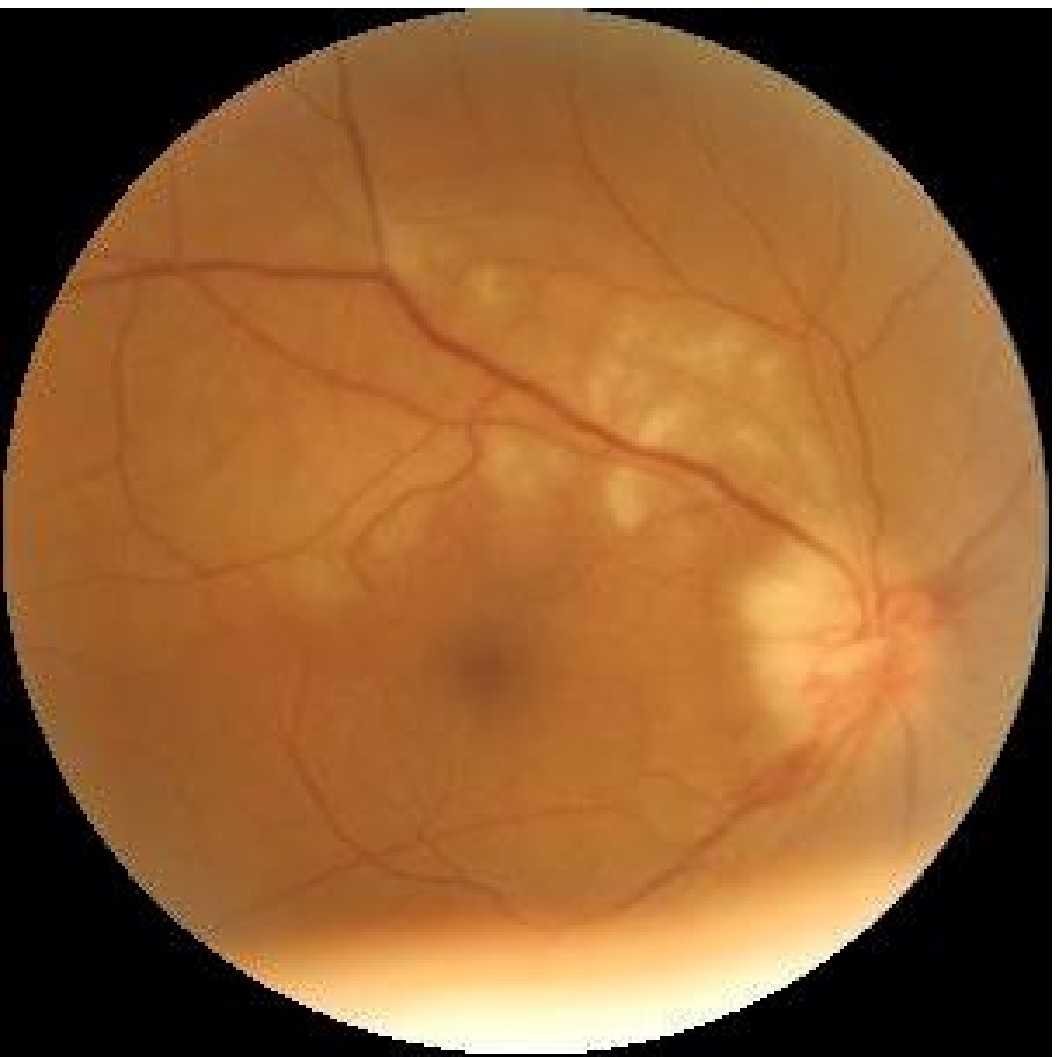}
\includegraphics[width=0.5in]{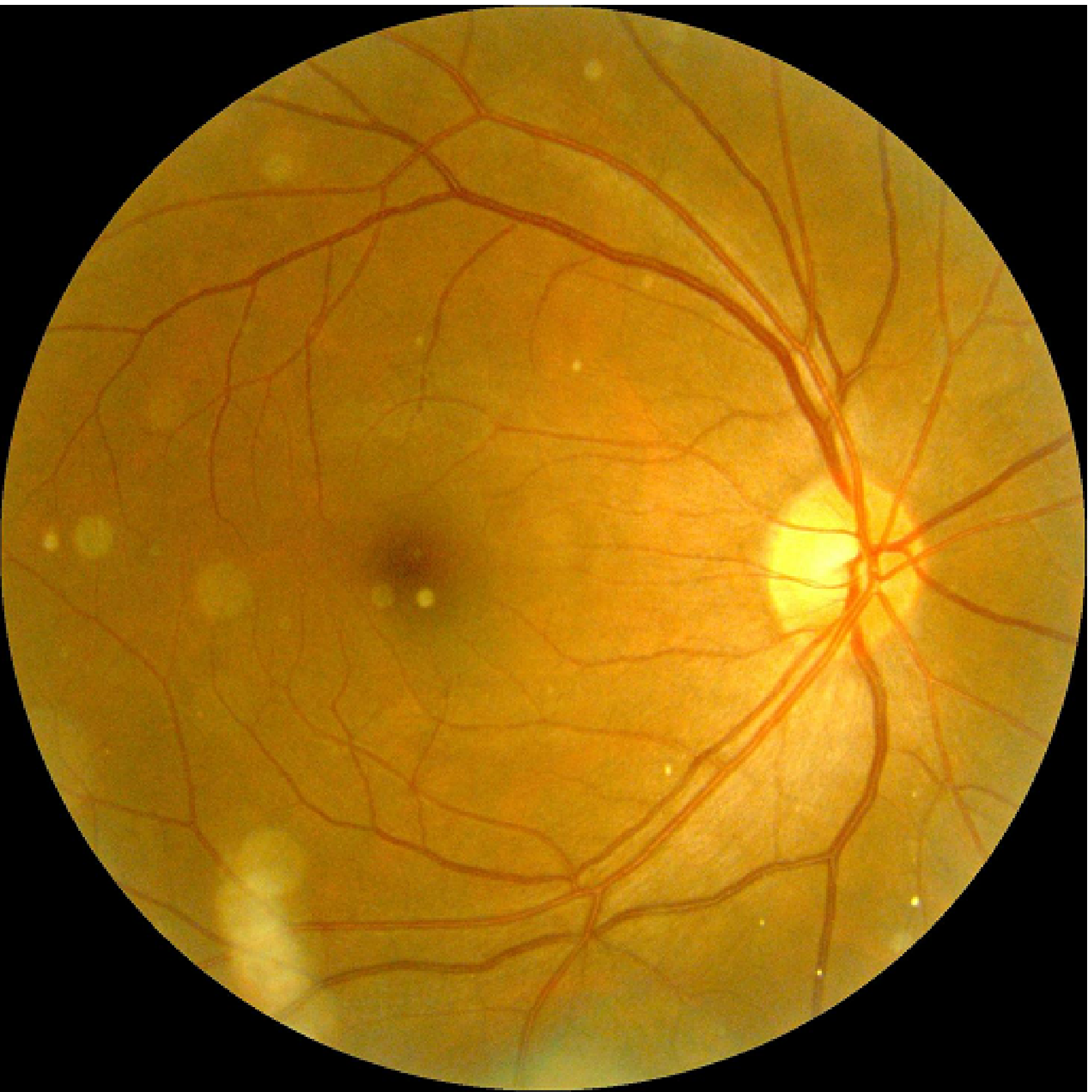}
\includegraphics[width=0.5in]{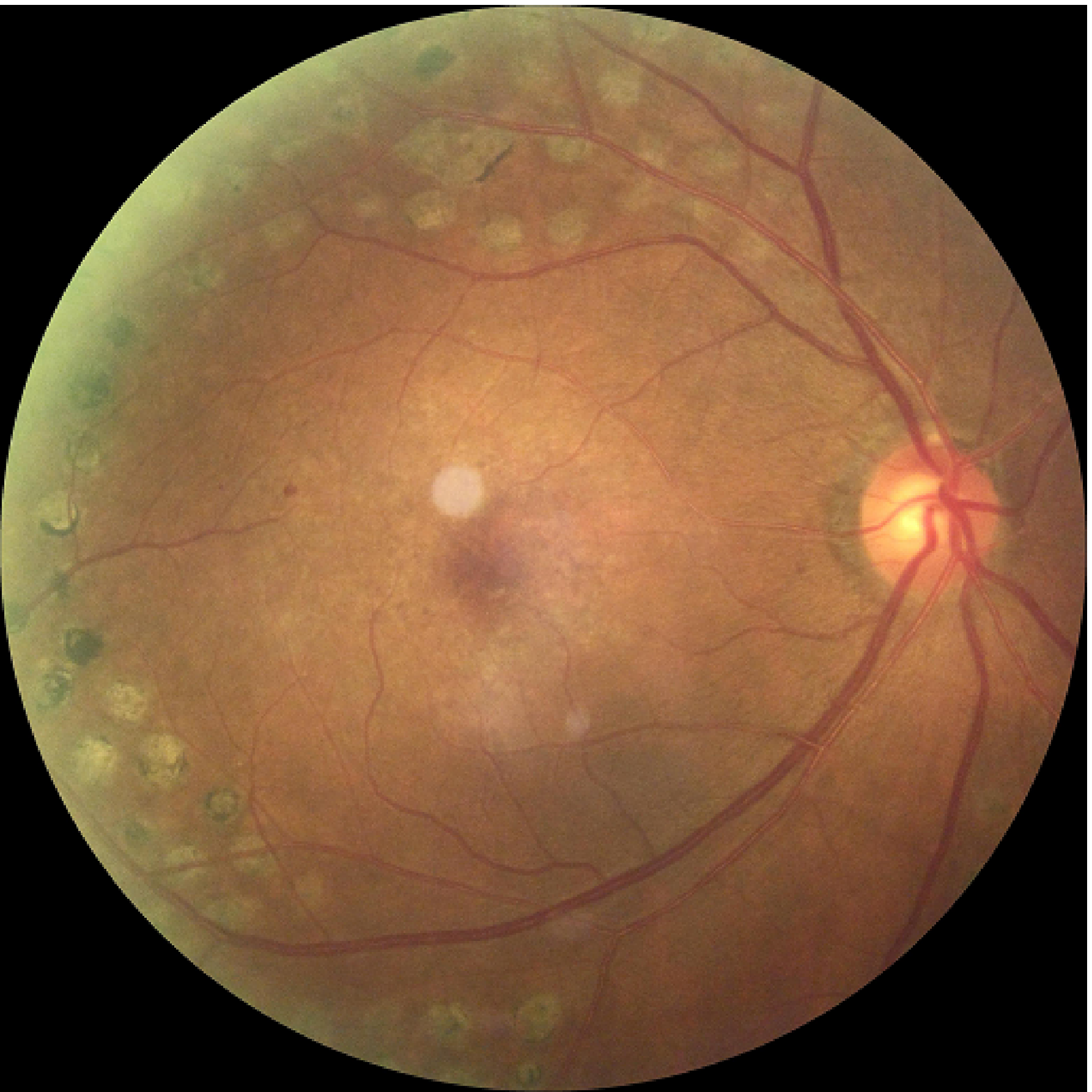}
\includegraphics[width=0.5in]{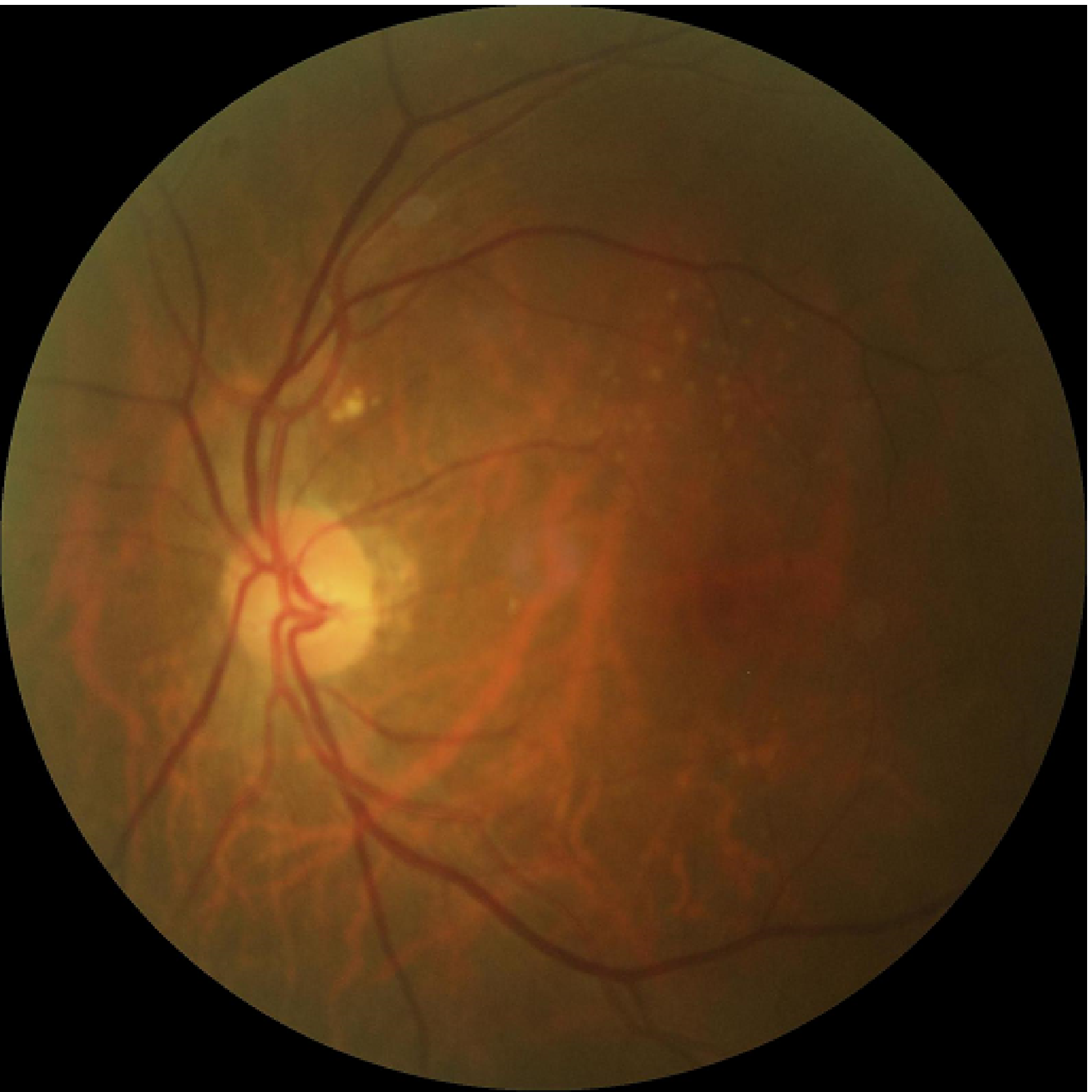}
\end{minipage}
}
\centering
\caption{Images from each categories. Each column of images comes from the same category, i.e. Normal (N), DR (D), Glaucoma (G), Cataract (C), AMD (A), Hypertension (H), Myopia (M), and Others (O)}
\label{fig3}
\end{figure}

\item \textit{Binocular-based:} Most of the existing ophthalmic disease detection works are based on a fundus image, but in real clinical scenarios, ophthalmologists usually diagnose patients with information from both eyes. In order that the related work performed on our dataset can be better applied to realistic scenes, our dataset contains fundus images of the left and right eyes of patients. Compared with detection on one eye merely, screening for patients with both eyes is both comprehensive and complicated, because in the process of feature extraction of the image, we have to balance the correlation between the two eyes and their respective characteristics. This makes the classification task on this dataset full of challenges. The ultimate goal of our dataset is to perform multi-label classification of patients using fundus images of both eyes. Fig.~\ref{fig2} shows some images of the patient's left and right eyes.

\item \textit{Multi-modal data:} Our dataset integrates multiple information of patients. In addition to providing researchers with fundus images of the left and right eyes of patients, we also provide age and gender of each patient and the diagnostic keywords of ophthalmologists for each fundus image, as shown in Fig.~\ref{fig2}. These information can help researchers better perform multi-disease classification tasks, and can also help them perform some more detailed research based on the dataset, such as generating textual diagnosis~\cite{chelaramani2019multi} and age predicting~\cite{poplin2018prediction} based on fundus images.

\item \textit{Scale:} In today's data-driven deep learning research, large-scale datasets are the cornerstone of ensuring that a research effort is truly applied to real-world scenarios. Our dataset contains 10,000 fundus images from the left and right eyes of 5,000 clinical patients. These images were acquired by different cameras at multiple ophthalmic centers in various regions of China. With a variety of resolutions and covering large scenes,
    we hope that it can promote further development in this field.
\end{enumerate}

\subsection{Split of Dataset}
The OIA-ODIR dataset consists of 10,000 fundus images from 5,000 clinical patients.
To evaluate the computer-aided algorithms of ocular diseases recognition, we split the dataset into three parts, i.e. the training set, the off-site test set and the on-site test set, which contains 3,500, 500 and 1,000 patients, respectively. The proportion of the training set and the test set is 7:3. The training set is used for training deep networks, and the off-site test set could be used as the validation set for model selection. The generalization ability of the deep network is evaluated on the on-site test set.

\setlength{\tabcolsep}{4pt}
\begin{table}
\caption{Proportion of images per category in training and testing datasets}
\label{tab2}
\centering
\begin{tabular}{c|c c c c c c c c}
\hline
Labels & N & D & G & C & A & H & M & O\\
\hline
Training case &  1138 & 1130 & 215 & 212 & 164 & 103 & 174 & 982\\
Off-site Testing cases & 162 & 163 & 32 & 31 & 25 & 16 & 23 & 136\\
On-site Testing cases &  324 & 327 & 58 & 65 & 49 & 30 & 46 & 275\\
All Cases &  1624 & 1620 & 305 & 308 & 238 & 149 & 243 &1393\\
\hline
\end{tabular}
\end{table}

In detail, the proportion of images per category in the training set and test set is summarized in Table~\ref{tab2}. We can observe that there exists serious class imbalance in OIA-ODIR. Specially, the number of fundus images with hypertension (H) is less than one tenth of normal images (N), which is challenging for multi-label diseases recognition.

\section{Multi-Disease Classification}
In order to establish benchmark performance on our proposed OIA-ODIR dataset and  evaluate the performance of some popular deep classification networks.
In this section, we select nine currently popular deep convolutional neural networks, including Vgg-16, ResNet-18, ResNet-50, ResNeXt-50, SE-ResNet-50, SE-ResNeXt-50, Inception-v4, Densenet and CaffeNet. Their performance will be described below.

\subsection{Network Structure}
\begin{figure}
\centering
\includegraphics[width=0.8\textwidth,height=0.35\textwidth]{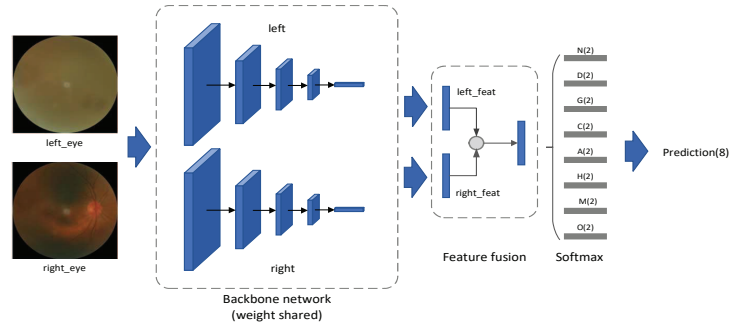}
\caption{The structure of multi-label classification network, the input of the network is two fundus images, and the backbone network is used for extracting their features. The feature fusion module fuses two features of left\_eye and right\_eye into one, and it is further fed into eight classifiers}
\label{fig4}
\end{figure}

In order to adapt to the characteristics of our OIA-ODIR, we made two modifications on the network structure.
Firstly, the input of the network is two images, rather than one.
Different from traditional image datasets used for diseases screening,
the status of patients take into account both left eyes and right eyes at the same time. Therefore, the inputs of the network are two fundus images and the corresponding ground-truth.
Furthermore, in order to better fuse the deep information of two fundus images and find a better baseline feature fusion method, we evaluated three feature fusion methods, including element-wise multiplication, element-wise sum and concat.
Secondly, it is multi-label classification task rather than single-label.
We added eight classifiers for two-class classification behind the last fully-connected layer of each network to achieve the purpose of multi-label classification.
Fig.~\ref{fig4} shows some details of our modified network.

\subsection{Experimental Settings}
\subsubsection{Running Environment}
All the deep networks, pre-trained on ImageNet, were trained and tested based on a publicly available convolutional network framework Caffe~\cite{jia2014caffe}. And, Caffe was compiled with CUDA 8.0 and CUDNN 5.1. The experiments ran on a workstation equipped with three NVIDIA GTX 1080ti GPUs, Intel E5 CPUs and 128GB memory, and the running operating system is Ubuntu 16.04.

\setlength{\tabcolsep}{2.5pt}
\begin{table}
\caption{The experimental results of nine deep networks on the Off-site and On-site testsets (Final denotes Final-score)}
\label{tab3}
\centering
\begin{tabular}{c | c | c c c c | c c c c}
\hline
\multirow{2}*{Fusion} & \multirow{2}*{Model} & \multicolumn{4}{c|}{Off-site} & \multicolumn{4}{c}{On-site}\\
\cline{3-10} && Kappa & F1 & AUC & Final & Kappa & F1 & AUC & Final \\
\hline
\multirow{9}*{SUM}
& \textcolor{red}{Vgg-16} & \textcolor{red}{0.4494} & \textcolor{red}{0.8730} & \textcolor{red}{0.8681} & \textcolor{red}{0.7302} & \textcolor{red}{0.4397} & \textcolor{red}{0.8718} & \textcolor{red}{0.8705} & \textcolor{red}{0.7273}\\

& ResNet-18 & 0.4325 & 0.8635 & 0.8422 & 0.7128 & 0.4137 & 0.8616 & 0.8365 & 0.7039\\

& ResNet-50 & 0.3799 & 0.8452 & 0.7988 & 0.6746 & 0.3827 & 0.8461 & 0.7885 & 0.6724\\

&\textcolor{blue}{ResNeXt-50} & \textcolor{blue}{0.4588} & \textcolor{blue}{0.8660} & \textcolor{blue}{0.8390} & \textcolor{blue}{0.7213} & \textcolor{blue}{0.4654} & \textcolor{blue}{0.8673} & \textcolor{blue}{0.8386} & \textcolor{blue}{0.7238}\\

& SE-ResNet-50 & 0.4140 & 0.8605 & 0.8693 & 0.7146 & 0.4265 & 0.8641 & 0.8689 & 0.7198\\

& SE-ResNeXt-50 & 0.4270 & 0.8660 & 0.8785 & 0.7238 & 0.4220 & 0.8666 & 0.8775 & 0.7220\\

&\textcolor{green}{Inception-v4} & \textcolor{green}{0.4507} & \textcolor{green}{0.8593} & \textcolor{green}{0.8800} & \textcolor{green}{0.7300} & \textcolor{green}{0.4487} & \textcolor{green}{0.8583} & \textcolor{green}{0.8669} & \textcolor{green}{0.7246}\\

& Densenet & 0.3914 & 0.8383 & 0.8472 & 0.6923 & 0.3971 & 0.8394 & 0.8460 & 0.6942\\

& CaffeNet & 0.3885 & 0.8563 & 0.8322 & 0.6923 & 0.3493 & 0.8460 & 0.8293 & 0.6749\\
\hline
\multirow{9}*{PROD}
&\textcolor{blue}{Vgg-16} &\textcolor{blue}{0.4359} &\textcolor{blue}{0.8665} &\textcolor{blue}{0.8545} &\textcolor{blue}{0.7190} &\textcolor{blue}{0.4527} &\textcolor{blue}{0.8700} &\textcolor{blue}{0.8628} &\textcolor{blue}{0.7284}\\
&ResNet-18 & 0.3593 & 0.8520 & 0.8493 & 0.6869 & 0.3798 & 0.8571 & 0.8583 & 0.6984\\
&ResNet-50 & 0.3545 & 0.8483 & 0.8372 & 0.6800 & 0.3697 & 0.8535 & 0.8408 & 0.6880\\
&\textcolor{green}{ResNeXt-50} &\textcolor{green}{0.4604} &\textcolor{green}{0.8660} &\textcolor{green}{0.8578}
&\textcolor{green}{0.7280} &\textcolor{green}{0.4626} &\textcolor{green}{0.8674} &\textcolor{green}{0.8499} &\textcolor{green}{0.7266}\\
&SE-ResNet-50 & 0.4321 & 0.8640 & 0.8613 & 0.7191 & 0.4096 & 0.8601 & 0.8571 & 0.7090\\
&SE-ResNeXt-50 & 0.4224 & 0.8663 & 0.8711 & 0.7199 & 0.4033 & 0.8630 & 0.8635 & 0.7099\\
&\textcolor{red}{Inception-v4} & \textcolor{red}{0.5063} & \textcolor{red}{0.8793} & \textcolor{red}{0.8691}
&\textcolor{red}{0.7516} &\textcolor{red}{0.4505} & \textcolor{red}{0.8668} &\textcolor{red}{0.8363} &\textcolor{red}{0.7178}\\
&Densenet & 0.4187 & 0.8415 & 0.8142 & 0.6915 & 0.3977 & 0.8338 & 0.7972 & 0.6762\\
&CaffeNet & 0.3678 & 0.8535 & 0.8495 & 0.6903 & 0.3531 & 0.8525 & 0.8466 & 0.6841\\
\hline
\multirow{9}*{Concat}
&\textcolor{green}{Vgg-16} &\textcolor{green}{0.3914} &\textcolor{green}{0.8658} &\textcolor{green}{0.8806} &\textcolor{green}{0.7126} &\textcolor{green}{0.3808} &\textcolor{green}{0.8641} &\textcolor{green}{0.8719} &\textcolor{green}{0.7056}\\
&ResNet-18 & 0.3299 & 0.8400 & 0.8480 & 0.6727 & 0.3674 & 0.8485 & 0.8488 & 0.6882\\
&ResNet-50 & 0.3421 & 0.8350 & 0.7853 & 0.6541 & 0.3292 & 0.8320 & 0.7928 & 0.6513\\
&ResNeXt-50 & 0.3568 & 0.8605 & 0.8523 & 0.6899 & 0.3383 & 0.8574 & 0.8477 & 0.6811\\
&\textcolor{blue}{SE-ResNet-50} &\textcolor{blue}{0.3940} &\textcolor{blue}{0.8660} &\textcolor{blue}{0.8702}
&\textcolor{blue}{0.7101} &\textcolor{blue}{0.3707} &\textcolor{blue}{0.8618} &\textcolor{blue}{0.8600} &\textcolor{blue}{0.6975}\\
&\textcolor{red}{SE-ResNeXt-50} &\textcolor{red}{0.4179} &\textcolor{red}{0.8593} &\textcolor{red}{0.8593}
&\textcolor{red}{0.7121} &\textcolor{red}{0.4091} &\textcolor{red}{0.8581} &\textcolor{red}{0.8606} &\textcolor{red}{0.7093}\\
&Inception-v4 & 0.3737 & 0.8500 & 0.8475 & 0.6904 & 0.3868 & 0.8518 & 0.8499 & 0.6961\\
&Densenet & 0.3072 & 0.8495 & 0.8306 & 0.6624 & 0.2772 & 0.8438 & 0.8211 & 0.6473\\
&CaffeNet & 0.3412 & 0.8485 & 0.8388 & 0.6762 & 0.3467 & 0.8500 & 0.8399 & 0.6789\\
\hline
\end{tabular}
\end{table}

\subsubsection{Evaluation Metrics}
We use four evaluation metrics, including Kappa, F1-score (F1), AUC and their mean value, denoted as Final-score, to evaluate the performance of multi-label classification networks.
Kappa coefficient is used for consistency check, and it ranges from -1 to 1.
F1 is the harmonic mean of precision and recall, which is high only when precision and recall are both high.
Since Kappa and F1 only consider a single threshold, while the output of classification networks is probabilistic, so that we use the area under the ROC curve (AUC) to comprehensively consider multiple thresholds.
All these four metrics are calculated by sklearn package.

\subsection{Experiment Analysis}
Nine deep networks were evaluated separately on two test sets, i.e. Off-site and On-site. Meanwhile, we experimentally verified three common feature fusion methods. Table~\ref{tab3} shows the results of our experiments using three different feature fusion methods. In the table we mark the top three results of each feature fusion method with red, green, and blue respectively. We find that compared with the other two feature fusion methods, the experiments using element-wise sum (SUM) feature fusion method gets better comprehensive performance.

\subsubsection{Experimental Results}
We have performed experiments on our dataset for convolutional neural network structures of different depths and widths.
To verify our conjecture, we analyzed the results of each group of experiments on the Off-site and On-site testsets, as shown in Table~\ref{tab3}.

\begin{figure}
\centering
\includegraphics[width=0.95\textwidth,height=0.45\textwidth]{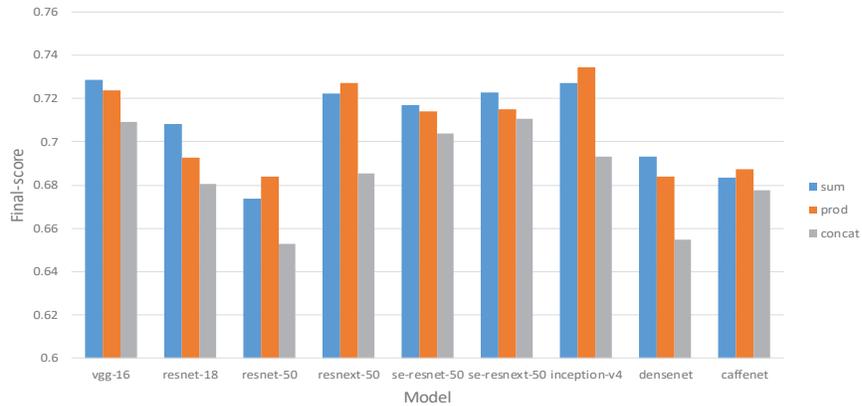}
\caption{Comparison of experimental performance among different models (sum denotes element-wise sum, prod denotes element-wise multiplication, Final-score is the mean value on the Off-site and On-site testsets)}
\label{fig5}
\end{figure}

In terms of network depth, Vgg-16 shows better performance than other deeper convolutional networks, so we think that simply increasing the depth of the neural network will not bring better results for our task.
Correspondingly, in terms of network width, such as ResNeXt-50 and Inception-v4, both have some similar characteristics, they increase the width of the neural network to combine the features of different dimensions to obtain a better result, which is similar with the characteristics of multiple diseases on one eye.
At the same time, by introducing the attention mechanism, the SE module can achieve certain effects under certain conditions, such as resnet-50, resnext-50 compared to se-resnet-50, se-resnext-50 in Concat feature fusion mode.
In addition, through the distribution of each category on the test set, we found that our samples are imbalance seriously, as shown in Table~\ref{tab2}, which also poses a huge challenge to our task. For this issue, we may need more labeled data and sampling methods to support our mission.

On the other hand, through the experimental verification of the three feature fusion methods, as shown in Fig.~\ref{fig5}, we find that the results of the two feature fusion methods of element-wise sum and element-wise multiplication are similar on each network model, which further illustrates the correctness of our conclusion above.
Meanwhile, for concat feature fusion method, we find that the distribution of evaluation indicators is different from the other two sets of experiments. However, because the evaluation results are not as good as other feature fusion methods, we think that concatenating the features of the left and right eyes simply can't bring a good improvement to our task, and its reference meaning is not significant.
In view of the above problems, we believe that a more structured feature fusion method is needed.

\begin{figure}
\centering
\includegraphics[width=0.95\textwidth,height=0.5\textwidth]{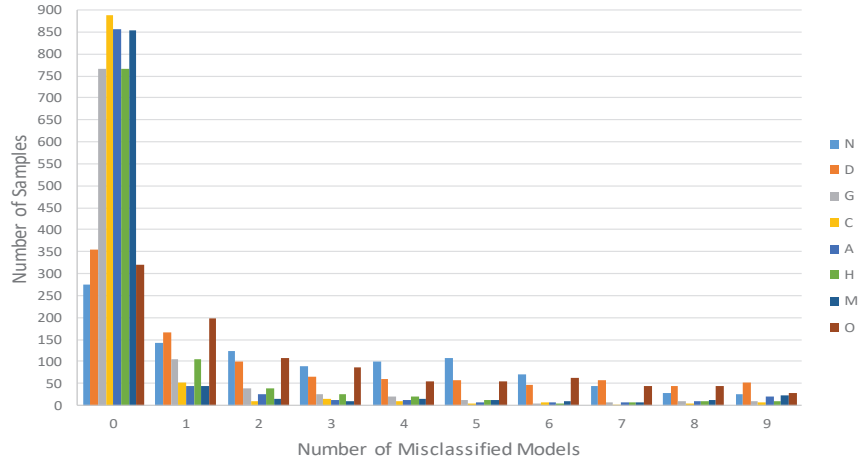}
\caption{Statistics on the number of samples with single label misclassified by n (0-9) models}
\label{fig7}
\end{figure}

\subsubsection{Bad Case Analysis}
In order to learn the classification of image samples, on the On-site testset with SUM feature fusion method, we performed statistics on the number of samples that were misclassified by the number of models on each label. Fig.~\ref{fig7} shows the details of our statistics. We can find that the analysis of single label is more helpful for our classification task, and most of the samples of each label can be correctly classified by at least 5 models. If the results of multiple models can be effectively integrated in the process of classifier design, we believe that better multi-label classification results will be achieved.

\begin{figure}
\centering
\includegraphics[width=0.95\textwidth]{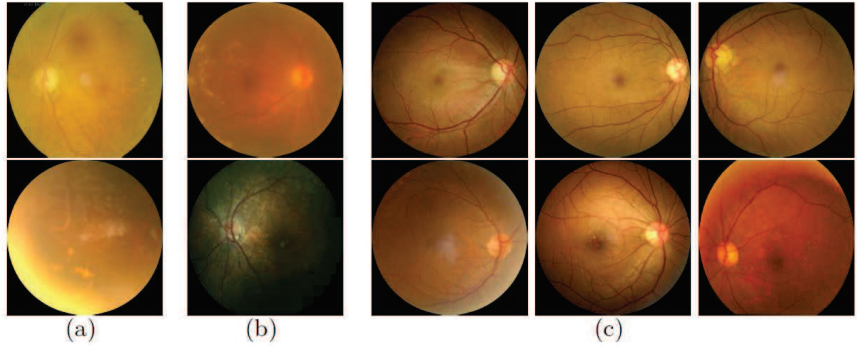}
\caption{Examples of misclassified images owing to different reasons. (a) camera lighting and lens stains, (b) multiple diseases interfere with each other, (c) local features are not obvious}
\label{fig8}
\end{figure}

We show some typical samples which are misclassified by most models in Fig.~\ref{fig8}. These images reflect some challenges and deficiencies of our dataset.
We summarize three key points.
1)~\textit{Image quality}. The images in (a) belong to diabetic retinopathy, in which hard exudate can be seen clearly. However, because of some lens stains and lighting problems, these images are misclassified.
2)~\textit{Intra-class confusion}. The classification of images in (b) are affected and confused by various fundus abnormalities. For example, cataract prevents the model from identifying hard exudate, hard exudation and drusen are difficult to be distinguish from each other. Which makes most of our misclassified samples are difficult to be extracted valid regional features by models.
3)~\textit{Local features are not obvious}. As shown in the images of the first line in (c), the determination of glaucoma requires accurate ratio of optic cup and disc. In the second line of (c), AMD requires more detailed features of the macular area. For these, we need to extract relevant local features to improve the performance of classification.


\section{Discussion}
As one publicly available multi-disease recognition dataset,
we have discussed the characteristics of our dataset.
In this section, we will discuss its challenges, deficiencies and potential applications.

\paragraph{\textit{Challenge:}} First of all, unlike previous studies of medical image processing based on a single fundus image, feature learning using left and right eye fundus images as input requires more considerations, such as the spatial and structural correlation of the two eyes. Second, for the classification of multiple disease labels, different fundus diseases have different local or global characteristics. For example, glaucoma is characterized by the relative size of the optic cup and optic disc, and cataracts are reflected in the clarity of various structures of the fundus. How to make suitable feature extraction schemes for different diseases in model design is a challenging task. And when multiple diseases interact in one fundus image, as shown in Fig.~\ref{fig8} (b), it is very difficult to identify them all correctly. Third, for the image itself, as shown in Fig.~\ref{fig3}, due to the wide source of our images, there is a wealth of intra-class diversity. In Fig.~\ref{fig3} (a), although they are all labeled as normal fundus, there are large differences in color, lighting, and lens shooting conditions, and the same situation also exists in other categories. At the same time, because the images in Other category are composed of a variety of uncommon fundus diseases, the differences within the class are more serious.

\paragraph{\textit{Potential Application:}} On one hand, our dataset can be used as a multi-disease screening dataset. On the other hand, there are several potential applications of our dataset.
1) We can explore the correlation between fundus images and patient's age or gender~\cite{poplin2018prediction}. Specifically, we can train a deep neural network for age predicting, and use a heat map to visualize the decision basis of the network.
2) In clinical scenarios, it is not enough to provide only classification results, the textual description of the fundus images will be more helpful~\cite{chelaramani2019multi}. Our dataset facilitates the development of image caption algorithms since diagnostic keywords of ophthalmologists for each fundus image are provided.
3) Multi-model data makes it possible to develop more accurate disease identification models. For example, graph neural network can be used to integrate un-structured data to further improve diagnostic accuracy.

\paragraph{\textit{Deficiency:}} As a work to make a bold attempt to detect multiple fundus diseases in a single fundus, our dataset also has some limitations. As a fundus image dataset with real clinical applications as the background, we believe that just 10,000 images can not really meet our application needs. In order to make related work safer and more accurate for clinical diagnosis, we need more fundus images. This can make our trained model have good generalization ability. At the same time, it can be found from Table~\ref{tab2} that there are serious data imbalances in different categories of our dataset. This is because some fundus diseases rarely occur clinically, and because multiple labels need to be labeled on a single image, it is difficult to ensure a balanced ratio between each category. In addition, although our dataset already provided detailed diagnostic keywords for each fundus image, they were finally divided into 8 categories. In future work, we need to perform some more fine-grained classification of fundus diseases. Finally, as an international fundus dataset, the source of images is only limited to some regions in China. In order to have a better geographical diversity of the dataset, we need fundus data from different races around the world.

\section{Conclusions}
Due to the lack of benchmark dataset hindering further development in automatic classification of clinical fundus images, in this paper, we release a fundus image dataset with multi-disease annotations.
The dataset contains 10,000 images from the left and right eyes of 5,000 clinical patients, with rich diversity.
At the same time, we evaluate the performance of some existing deep learning models on our dataset, in order to provide a valuable reference for future related work.
Experimental results show that simply increasing the depth of the neural network alone cannot improve performance, however increasing the width of the network can bring certain improvement. Moreover, integrating multiple deep networks will be helpful to improve classification performance.
At last, the dataset is available at \url{https://github.com/nkicsl/OIA-ODIR}.

\section*{Acknowledgements}
This work is partially supported by the National Natural Science Foundation (61872200), the Open Project Fund of State Key Laboratory of Computer Architecture, Institute of Computing Technology, Chinese Academy of Sciences No. CARCH201905, the Natural Science Foundation of Tianjin (19JCZDJC31600, 18YFYZCG00060).

 \bibliographystyle{splncs04}
 \bibliography{odir_ben20}

%
%
%
%
\end{document}